\documentclass{article}

 \usepackage[eandd, final]{neurips_2026}
\usepackage[utf8]{inputenc} % allow utf-8 input
\usepackage[T1]{fontenc}    % use 8-bit T1 fonts
\usepackage{hyperref}       % hyperlinks
\usepackage{url}            % simple URL typesetting
\usepackage{booktabs}       % professional-quality tables
\usepackage{amsfonts}       % blackboard math symbols
\usepackage{nicefrac}       % compact symbols for 1/2, etc.
\usepackage{microtype}      % microtypography
\usepackage{xcolor}         % colors
\usepackage{graphicx}
\usepackage{enumitem}
\usepackage{multicol}
\usepackage{multirow}
\usepackage{subcaption}
\usepackage{makecell}

\title{ECG-Reasoning-Benchmark: A Benchmark for Evaluating Clinical Reasoning Capabilities in ECG Interpretation}

\author{%
  Jungwoo Oh \\
  KAIST \\
  \texttt{ojw0123@kaist.ac.kr} \\
  \And
  Hyunseung Chung \\
  KAIST \\
  \texttt{hs\_chung@kaist.ac.kr} \\
  \And
  Junhee Lee \\
  KAIST \\
  \texttt{ciel3486@kaist.ac.kr} \\
  \And
  Min-Gyu Kim \\
  Ajou University School of Medicine \\
  \texttt{manjmin6@gmail.com} \\
  \And
  Hangyul Yoon \\
  KAIST \\
  \texttt{hangyulmd@kaist.ac.kr} \\
  \And
  Ki Seong Lee \\
  KAIST \\
  \texttt{kistar12@kaist.ac.kr} \\
  \And
  Youngchae Lee \\
  Yonsei University College of Medicine \\
  \texttt{dudco5307@yuhs.ac} \\
  \And
  Muhan Yeo \\
  Seoul National University Bundang Hospital \\
  \texttt{smfforsm@gmail.com} \\
  \And
  Edward Choi \\
  KAIST \\
  \texttt{edwardchoi@kaist.ac.kr} \\
}

\begin{document}

\maketitle

\begin{abstract}
    While Multimodal Large Language Models (MLLMs) show promising performance in automated electrocardiogram interpretation, it remains unclear whether they perform actual step-by-step reasoning or just rely on superficial visual cues.
    To investigate this, we introduce \textbf{ECG-Reasoning-Benchmark}, a novel multi-turn evaluation framework comprising over 6,400 samples to systematically assess step-by-step reasoning across 17 core ECG diagnoses.
    Our comprehensive evaluation of state-of-the-art models reveals a systematic shortfall in executing multi-step logical deduction.
    Although models possess the medical knowledge to retrieve clinical criteria for a diagnosis, they exhibit consistently low success rates in maintaining a complete reasoning chain, primarily failing to ground the corresponding ECG findings to the actual visual evidence in the ECG signal.
    These results demonstrate that current MLLMs bypass actual visual interpretation, highlighting a limitation in their visual grounding capabilities and motivating robust, reasoning-centric medical AI. The code and data are available at \url{https://github.com/Jwoo5/ecg-reasoning-benchmark}.
\end{abstract}

\section{Introduction}

Over the past decade, deep learning has revolutionized automated electrocardiogram (ECG) interpretation.
Discriminative models have achieved diagnostic accuracy comparable to, and occasionally surpassing, human cardiologists in classification tasks~\cite{Pyakillya_2017, LIU2021107187}.
However, clinical practice remains hesitant to adopt these models.
In the high-stakes domain of healthcare, a ``black-box'' prediction is insufficient.
Clinicians require not just a diagnostic label, but the clinical reasoning and evidence that justify the conclusion to make informed decisions.

To bridge this interpretability gap, the field has rapidly pivoted toward Multimodal Large Language Models (MLLMs).
Recent works such as PULSE~\cite{liu2024teach}, GEM~\cite{lan2025gem}, OpenTSLM~\cite{langer2025opentslm}, and ECG-R1~\cite{jin2026ecg} integrate ECG signals with Large Language Models (LLMs) to generate diagnostic reports or answer clinical queries.
While these models can generate fluent and plausible-sounding explanations, they introduce a new and dangerous risk: hallucination.
% This risk of hallucination stems primarily from how their training data is constructed.
In many existing datasets, the training explanations are synthetically generated by providing an LLM like GPT-4~\cite{achiam2023gpt} with the final diagnostic labels and machine-generated reports, typically without direct exposure to the actual ECG signal.
Because the models are trained on these text-derived rationales rather than visually grounded features, they often struggle to ground their interpretations to raw physiological evidence.
Instead, they learn to generate medically fluent justifications that recite textbook descriptions associated with the diagnosis, regardless of what the underlying signal actually shows.

Furthermore, the prevailing evaluation methodology worsens this disconnect.
Existing studies predominantly rely on the LLM-as-a-Judge framework~\cite{zheng2023judging}, which evaluates the generated interpretation by comparing it against a reference response.
A major limitation of this approach is that these reference explanations are also synthetically created by an LLM.
Therefore, measuring the alignment between a model's output and these references primarily assesses how well the model mimics the linguistic style of the text generator.
Because the judge-LLM never looks at the actual ECG image to verify the model's response, this framework can validate whether an explanation is medically plausible and fluent, but cannot verify if the interpretation is grounded in the underlying ECG signal.

To address these limitations in current evaluation paradigms, we propose \textbf{ECG-Reasoning-Benchmark}.
We posit that evaluating an ECG-MLLM should not be a test of fluency, but a rigorous ``clinical reasoning exam'' that probes the model's intrinsic reasoning capabilities.
We conceptualize ECG interpretation as a multi-stage deduction process requiring established medical knowledge, perceptual detection, and precise visual grounding of ECG features.
To rigorously assess this process, we implement a 4-stage verification loop that sequentially evaluates the reasoning trajectory from initial criterion selection to the final diagnostic decision.
% This framework ensures that a correct diagnosis is the result of a verifiable chain of evidence.

Our contributions are summarized as follows:
% \vspace{-3mm}
\begin{enumerate}
    \item We propose \textbf{ECG-Reasoning-Benchmark}, a novel evaluation framework grounded in established clinical criteria and precise ECG features. This shifts the evaluation paradigm from subjective LLM-as-a-Judge scoring to rigorous step-by-step verification, providing a reliable standard to ascertain whether models base their decisions on the actual ECG signal.
    \item To facilitate this rigorous evaluation, we develop a comprehensive automated ECG analysis pipeline that extracts explicit diagnostic features directly from raw 12-lead signals. By progressively mapping wave delineations and quantitative measurements to discrete clinical findings, this tool establishes a transparent and objective ground truth for the clinical reasoning chains required for our benchmark.
    \item Through a comprehensive evaluation of state-of-the-art MLLMs, we reveal a systematic gap in multi-step logical deduction. We demonstrate that while current models possess the medical knowledge to identify which ECG findings are required for a diagnosis, they lack the capability to ground those specific findings within the ECG signal, highlighting a current limitation in their grounding capabilities.
\end{enumerate}

\section{Related works}

The initial wave of ECG-MLLMs focused on adapting general-purpose vision-language architectures to the cardiac domain.
Early initiatives such as MEIT~\cite{wan2025meit}, ECG-LM~\cite{yang2025ecg}, and Q-Heart~\cite{pham2025q} treated ECG interpretation as a translation task, mapping global signal embeddings to clinical reports (\textit{i.e.,} report generation task) or text-based answers (\textit{i.e.,} question answering task).
While effective for these specific tasks, they still lack the capacity to explain the grounded evidence underlying their outputs.

Subsequent research attempted to bridge this gap by incorporating explicit reasoning processes into both models and datasets.
However, they rely on synthetic data generation processes.
For instance, PULSE~\cite{liu2024teach} is fine-tuned on the ECGInstruct dataset, where instruction-response pairs were synthesized by Llama-3-70B-Instruct~\cite{grattafiori2024llama} without direct exposure to actual signals.
Recognizing the need for structural grounding, later efforts attempted to incorporate physical measurements extracted from an external tool~\cite{hong2017encase,hong2019combining}.
GEM~\cite{lan2025gem} introduced the ECG-Grounding dataset, and ECG-R1~\cite{jin2026ecg} utilized the ECG-Protocol-Guided-Grounding-CoT dataset.
Meanwhile, OpenTSLM~\cite{langer2025opentslm} employed the ECG-QA-CoT dataset, which relies on Chain-of-Thought trajectories generated by GPT-4o~\cite{hurst2024gpt} from question-answer pairs in the ECG-QA dataset~\cite{oh2023ecg}.
Other approaches like ECG-Chat~\cite{zhao2025ecg} integrated Retrieval-Augmented Generation to mitigate hallucination.

Despite these advancements, a fundamental limitation persists in these methodologies due to their reliance on synthetic ground truth.
Because the reasoning chains in these works are synthesized by LLMs, trained models learn to emulate the linguistic style of the teacher model rather than deriving evidence from the raw signal.
Furthermore, the prevailing LLM-as-a-Judge evaluation frameworks cannot verify whether the interpretations are actually supported by the input signal.
To address these structural limitations, our \textbf{ECG-Reasoning-Benchmark} provides an objective and quantitative examination of explicit clinical reasoning grounded in the ECG signal.

\begin{figure}
    \centering
    \includegraphics[width=1.0\linewidth]{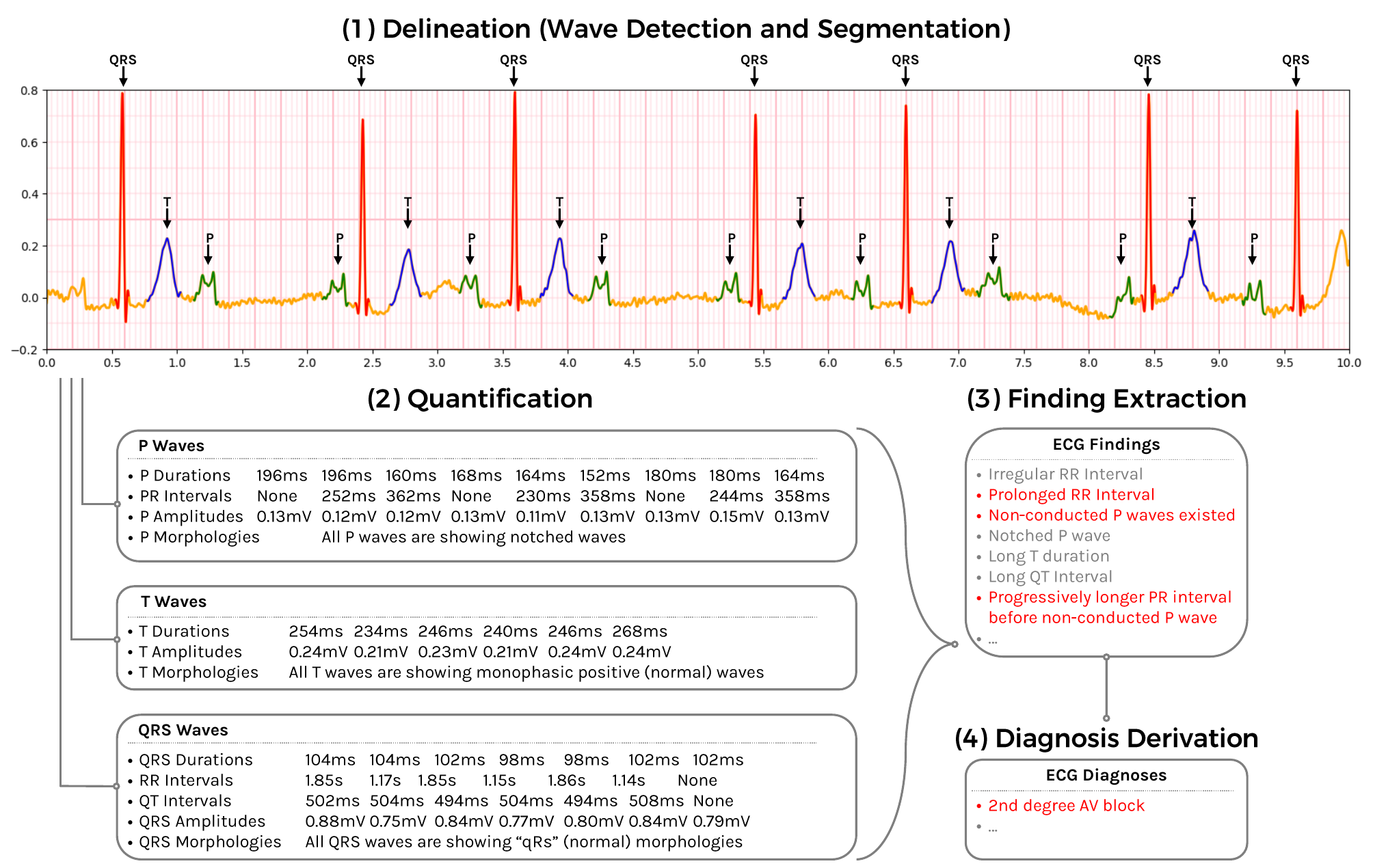}
    \caption{End-to-end schematic of the Automated ECG Analysis Pipeline. The workflow illustrates a systematic progression: from the initial delineation of fundamental waveforms in the raw signal to the quantification of physiological features, which are then mapped into discrete clinical findings and ultimately resulting in a definitive diagnostic conclusion.}
    \label{fig:pipeline}
    % \vspace{-2mm}
\end{figure}

\section{Automated ECG analysis pipeline}
\label{sec:pipeline}

To facilitate a rigorous reasoning benchmark, it is imperative to establish a ground truth that provides a transparent and traceable chain of clinical evidence.
However, such granular annotations are largely absent from existing public ECG datasets, which typically provide high-level diagnostic labels without the exact waveform boundaries or specific interval measurements required for clinical reasoning.
To overcome this, we developed an \textbf{Automated ECG Analysis Pipeline}, which constructs verifiable ground-truth annotations by systematically extracting physiological evidence directly from the raw signal.
% To provide a clear overview of this systematic extraction process,
The schematic illustration of this pipeline is provided in Figure~\ref{fig:pipeline}.

\subsection{Wave detection and segmentation}
\label{sec:segmentation}

The foundation of the pipeline lies in the precise delineation of the P wave, QRS complex, and T wave.
To achieve this, we employ a U-Net3+ architecture~\cite{joung2024deep} to perform the initial wave detection.
For a given 12-lead ECG, the model processes each lead individually, generating separate probability maps for four classes: P wave, QRS complex, T wave, and the isoelectric background.
To refine these initial outputs for clinical validity, we further apply context-aware post-processing algorithms comprising three components: (1) \textbf{P-wave recovery via template matching}, which performs a secondary template-guided search within RR intervals to recover non-conducted P waves that the deep model often misses (\textit{e.g.,} in high-degree AV blocks); (2) \textbf{Physiological constraint enforcement}, which applies strict biological rules such as enforcing a single T wave per cardiac cycle to eliminate artifacts; and (3) \textbf{Multi-lead consensus alignment}, which validates a wave only when it is detected at a consistent temporal location across at least 4 of the 12 leads (See Appendix~\ref{sec:app_postproc} for more details).

Given that precise wave delineation is the critical foundation for all subsequent ECG analysis, we evaluated the performance of this detection module.
Specifically, empirical evaluations on the Lobachevsky University Electrocardiography Database (LUDB)~\cite{kalyakulina2020ludb} indicate that our pipeline provides robust detection accuracy compared to traditional signal processing baselines.
The pipeline achieves an average recall and precision of 1.000 for QRS complexes, 0.978 and 0.937 for P waves, and 0.996 and 0.992 for T waves.
Detailed results are provided in Appendix~\ref{sec:app_seg_result}.

\subsection{Feature extraction}
\label{sec:feat_extr}

\paragraph{Quantification}
Following segmentation, the pipeline proceeds to a hierarchical feature extraction phase, starting from the quantification of low-level ECG features.
This includes temporal measurements such as the duration of P, QRS, and T waves, alongside physiological intervals like PR, and RR intervals.
Simultaneously, amplitude measurements are computed by measuring peak heights relative to the isoelectric line and quantifying ST-segment deviations at the J-point.
Additionally, the pipeline performs morphological analysis to identify specific QRS structural configurations (\textit{e.g.,} qR, rS, and rSR' patterns) and pathological Q waves, and computes the frontal plane electrical axis for each beat.

\paragraph{Finding extraction}
Once these continuous quantitative measurements are extracted, the pipeline proceeds to map them to ECG findings.
This step bridges the gap between raw signal processing and medical terminology by applying established clinical criteria.
For instance, continuous interval values are evaluated against standard physiological limits, where a PR interval exceeding 200 ms in the majority of detected beats is formally identified as a ``Prolonged PR interval''.
% This transformation converts the dense, high-dimensional feature space into a discrete set of interpretable clinical findings.

\paragraph{Diagnosis derivation}
Finally, the pipeline maps the identified findings to a clinical diagnosis using hierarchical logic diagrams covering 17 core ECG diagnoses. These diagrams were codified from established guidelines such as the ECG Core Curriculum~\cite{2023ecgcorecurriculum} and validated by three board-certified internal medicine specialists (See Appendix~\ref{sec:app_diagrams}).
This strict framework confirms a diagnosis only if a clinically valid combination of findings is present, thereby establishing a ground truth that explicitly defines the causal chain of findings required to rigorously verify the model’s reasoning process.

\section{Construction of ECG-Reasoning-Benchmark}

Leveraging the structured ground truth derived from our automated pipeline, we constructed \textbf{ECG-Reasoning-Benchmark}.
% Distinct from traditional QA datasets, and diverging from recent reliance on subjective LLM-as-a-Judge approaches, our framework provides an objective testbed that verifies whether each step in the entire chain of clinical deduction is grounded in physical signal evidence.
Distinct from recent reliance on subjective LLM-as-a-Judge approaches, our framework provides an objective testbed that verifies whether each step in the entire chain of clinical deduction is grounded in physical signal evidence.

\begin{figure}[t]
    \centering
    \includegraphics[width=1.0\linewidth]{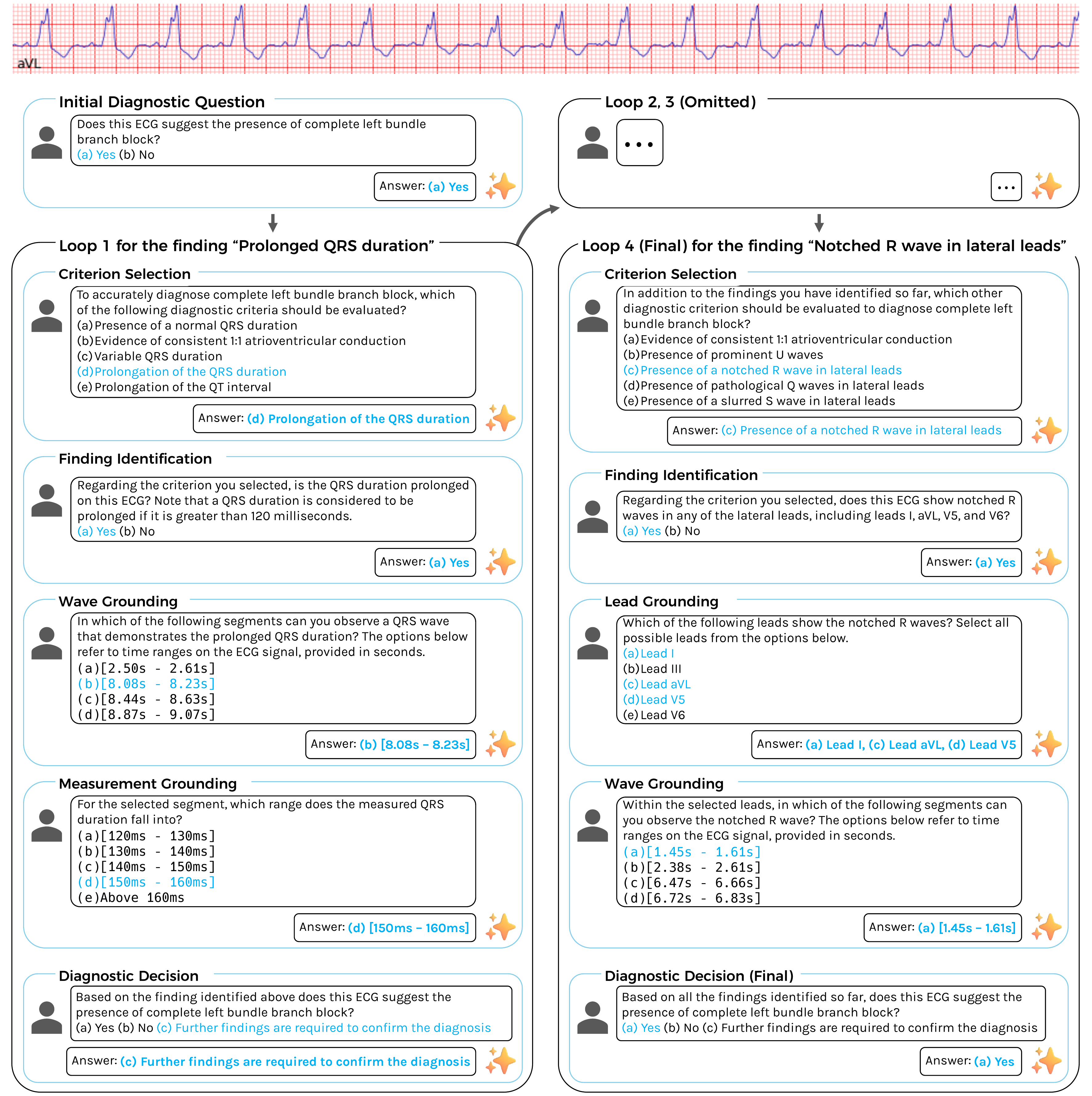}
    \caption{An illustrative example of the 4-step recursive evaluation loop for diagnosing Complete Left Bundle Branch Block. The process begins with an Initial Diagnostic Question, followed by the verification of necessary findings. The figure depicts the first loop, which verifies the presence of ``Prolonged QRS duration'' grounded in the wave segment and measurement range, and the fourth loop, which confirms the ``Notched R wave in lateral leads'' with specific lead and wave grounding.
    % The final diagnostic decision is reached only after all required findings are sequentially validated.
    }
    \label{fig:fig2}
    % \vspace{-2mm}
\end{figure}

\subsection{Evaluation workflow}

Inspired by the methodology of CXReasonBench~\cite{PhysioNet-chexstruct-cxreasonbench-1.0.1}, our evaluation protocol begins with an Initial Diagnostic Question (\textit{e.g., }``Does this ECG suggest the presence of first-degree AV block?'').
This step establishes a baseline for the model's intuitive diagnostic capability prior to engaging in detailed reasoning.
Regardless of whether the model answers this initial question correctly, the evaluation advances to the step-wise verification process, which systematically challenges the model to execute the chain of clinical deduction through four distinct steps applied to each individual clinical finding:
% \vspace{-2mm}
\begin{enumerate}[leftmargin=0.5cm]
    \item \textbf{Criterion Selection}: The model must first identify the specific diagnostic criterion relevant to the target diagnosis (\textit{e.g.,} ``To accurately diagnose complete left bundle branch block, which of the following diagnostic criteria should be evaluated?'').
    % To strictly evaluate discriminatory ability, we employ two types of distractors comprising of category-based and presence-based distractors.
    % Specifically, category-based distractors introduce incorrect options drawn from the same clinical category as the correct finding (\textit{e.g.,} contrasting ``Prolonged PR interval'' with ``Normal PR interval'').
    % On the other hand, presence-based distractors consist of findings that are present in the current ECG recording but are clinically irrelevant to the diagnosis in question.
    \item \textbf{Finding Identification}: Upon selecting a criterion, the model is challenged to verify its presence in the current recording (\textit{e.g.,} ``Is the QRS duration prolonged on this ECG?'').
    This assesses the model's fundamental perceptual capacity to detect abnormalities visually.
    \item \textbf{ECG Grounding}: To distinguish genuine analysis from hallucination, we demand explicit signal grounding, which involves three granular sub-tasks:
    \begin{itemize}[leftmargin=0.5cm]
        \item \textbf{Lead Grounding}: For findings associated with specific anatomical regions or lead groups, the model must identify the precise leads exhibiting the abnormality (\textit{e.g.,} ``Which of the following leads show the notched R waves?'').
        \item \textbf{Wave Grounding}: The model is required to temporally locate the relevant waveforms within the 10-second strip to demonstrate its visual focus (\textit{e.g.,} ``In which of the following segments can you observe the notched R wave?'').
        \item \textbf{Measurement Grounding}: The model quantifies the specific feature by selecting the correct value range (\textit{e.g.,} ``Which range does the measured QRS duration fall into?'')
    \end{itemize}
    To make the constrained response space explicit, we describe how grounding answer options and distractors are constructed in Appendix~\ref{sec:app_grounding_answer}.
    \item \textbf{Diagnostic Decision}: Finally, based on the verified findings, the model determines whether the diagnosis can be confirmed or if ``further findings are required'' (\textit{e.g.,} ``Based on all the findings identified so far, does this ECG suggest the presence of complete left bundle branch block?'').
\end{enumerate}

This 4-step validation sequence is applied iteratively for every clinical finding required to support the diagnosis, so the final diagnostic conclusion is reached only after every piece of supporting evidence has been explicitly validated. An example for diagnosing Complete Left Bundle Branch Block is visually provided in Figure~\ref{fig:fig2}, with the constituent criteria and their hierarchical arrangement derived from the verified logic diagrams described in Section~\ref{sec:feat_extr}.

\subsection{Sampling strategy}

To ensure the benchmark serves as a robust and unbiased evaluator of clinical reasoning, we implemented a sophisticated sampling strategy.
Specifically, we curated a balanced set of 100 positive and 100 negative samples for each of the 17 core diagnoses.
Crucially, since a single diagnosis can be confirmed through multiple combinations of clinical findings, we ensured that the selected samples were evenly distributed across the various logical paths defined in our logic diagrams.
This approach enables the benchmark to evaluate the model's competence across diverse clinical presentations.

To further guarantee the reliability of the ground truth, we strictly filtered the dataset to include only samples where the provided label aligns with our automated pipeline's diagnosis.
This verification process ensures that every sample is not only labeled in its source dataset but also quantitatively supported by verifiable signal features.
Applying this protocol to two source datasets, PTB-XL~\cite{wagner2020physionet} and MIMIC-IV-ECG~\cite{gowmimic}, we constructed a comprehensive benchmark comprising over 6,400 samples (3,084 from PTB-XL and 3,359 from MIMIC-IV-ECG), where the detailed dataset statistics can be found in Appendix~\ref{sec:app_data_stat}.
Finally, three board-certified internal medicine specialists validated 147 representative samples, one per unique reasoning path in each source dataset (73 from PTB-XL and 74 from MIMIC-IV-ECG).
The specialists examined the extracted reasoning path together with its grounding elements, namely the specific leads, wave segments, and measurement values, and reached a single consensus judgment per sample.
All remaining samples were manually reviewed by the authors under the direct supervision of the specialists, applying the same criteria to every grounding element.
Any sample that failed this review was removed and replaced by resampling from the same reasoning path, preserving the balance across paths.

\section{Experiments}

\subsection{Experimental setup}
\label{sec:expr_setup}

\paragraph{Evaluated models}
To comprehensively assess clinical reasoning capabilities, we evaluate a diverse suite of state-of-the-art Multimodal Large Language Models (MLLMs) processing either visual or time-series inputs.
Our evaluation includes ECG-specific models (PULSE~\cite{liu2024teach}, GEM~\cite{lan2025gem}, ECG-R1~\cite{jin2026ecg}, OpenTSLM~\cite{langer2025opentslm}), medical-domain models (Hulu-Med~\cite{jiang2025hulu}, MedGemma~\cite{sellergren2025medgemma}), open-weight general domain models (Qwen3-VL~\cite{bai2025qwen3}, Llama-3.2-Vision~\cite{grattafiori2024llama}), and proprietary models (Gemini-2.5-Flash~\cite{comanici2025gemini}, Gemini-2.5-Pro~\cite{comanici2025gemini}, Gemini-3-Flash~\cite{team2023gemini}, GPT-5-Mini~\cite{singh2025openai}, GPT-5.2~\cite{singh2025openai}).
Detailed model configurations are provided in Appendix~\ref{sec:app_impl_details}.

\paragraph{Data processing}
Depending on the architectural requirements of each model, the ECG data is inputted either as 1D time-series arrays or 2D images.
Specifically, for OpenTSLM, which is natively designed for time-series, the input is provided as a 100Hz, 12-channel time-series signal.
For all other vision-capable models, the 1D signals are converted into standard 12-lead 2D ECG images using the \texttt{ecg-plot} Python package.
Additionally, we provide a system prompt that instructs the models to strictly follow established clinical guidelines and complete the verification of all required findings before reaching a final diagnosis.
This system prompt is presented in Appendix~\ref{sec:prompt}.

\begin{table}[t]
    \centering
    \caption{Quantitative benchmark results of the evaluated models on \textbf{ECG-Reasoning-Benchmark} across both PTB-XL and MIMIC-IV-ECG source datasets. Models are grouped into ECG-specific, medical domain, general domain, and proprietary categories, with the highest score for each metric highlighted in bold within its respective source dataset.}
    \label{tab:result_main}
    \resizebox{\textwidth}{!}{
    \begin{tabular}{ccccccc}
        \toprule
        \multirow{3}{*}{\textbf{Source Dataset}} & \multirow{3}{*}{\textbf{Model}} & \multirow{3}{*}{\textbf{IDA (\%)}} & \multicolumn{2}{c}{\textbf{Completion (\%)}} & \multirow{3}{*}{\textbf{Depth (0-4)}} & \multirow{3}{*}{\textbf{GT-RDA (\%)}} \\
        \cmidrule(lr){4-5}
        & & & \textbf{Hard} & \textbf{Soft} & & \\
        \midrule
        \multirow{19}{*}{PTB-XL}& PULSE (7B)~\cite{liu2024teach} & $81.36$ & $3.89$ & $3.92$ & $1.24$ & $34.50$ \\
        & GEM (7B)~\cite{lan2025gem} & $84.40$ & $4.35$ & $5.19$ & $1.34$ & $67.28$ \\
        & ECG-R1-SFT (8B)~\cite{jin2026ecg} & $85.38$ & $5.90$ & $6.06$ & $1.79$ & $21.60$ \\
        & ECG-R1-RL (8B)~\cite{jin2026ecg} & \scalebox{1.1}{$\bf 85.86$} & $6.42$ & $6.68$ & $1.79$ & $22.60$ \\
        & OpenTSLM (3B)~\cite{langer2025opentslm} & $54.15$ & $0.71$ & $0.71$ & $0.18$ & $37.06$ \\
        \cmidrule{2-7}
        & Hulu-Med (7B)~\cite{jiang2025hulu} & $57.30$ & $4.54$ & $5.61$ & $1.24$ & $86.38$ \\
        & Hulu-Med (32B)~\cite{jiang2025hulu} & $57.13$ & $2.89$ & $3.76$ & $1.62$ & \scalebox{1.1}{$\bf 99.29$} \\
        & MedGemma (4B)~\cite{sellergren2025medgemma} & $51.43$ & $0.58$ & $0.75$ & $0.80$ & $32.85$ \\
        & MedGemma (27B)~\cite{sellergren2025medgemma} & $55.22$ & $1.23$ & $2.59$ & $1.52$ & $78.21$ \\
        & MedGemma-1.5 (4B)~\cite{sellergren2025medgemma} & $43.09$ & $0.62$ & $0.78$ & $0.92$ & $84.40$ \\
        \cmidrule{2-7}
        & Qwen3-VL (8B)~\cite{bai2025qwen3} & $53.70$ & $5.80$ & $6.23$ & $1.52$ & $87.48$ \\
        & Qwen3-VL (32B)~\cite{bai2025qwen3} & $59.40$ & $5.90$ & $7.00$ & $1.62$ & $89.53$ \\
        & Llama-3.2-Vision (11B)~\cite{grattafiori2024llama} & $46.69$ & $1.36$ & $1.75$ & $1.02$ & $78.79$ \\
        & Llama-3.2-Vision (90B)~\cite{grattafiori2024llama} & $56.23$ & $3.60$ & $4.90$ & $1.36$ & $91.02$ \\
        \cmidrule{2-7}
        & Gemini-2.5-Flash~\cite{comanici2025gemini} & $58.46$ & $5.29$ & $6.29$ & $1.84$ & $89.92$ \\
        & Gemini-2.5-Pro~\cite{comanici2025gemini} & $55.67$ & $3.08$ & $4.93$ & $1.99$ & $87.19$ \\
        & Gemini-3-Flash~\cite{team2023gemini} & $63.72$ & \scalebox{1.1}{$\bf 6.84$} & \scalebox{1.1}{$\bf 11.28$} & \scalebox{1.1}{$\bf 2.08$} & $91.96$ \\
        & GPT-5-Mini~\cite{singh2025openai} & $57.30$ & $2.11$ & $2.98$ & $1.75$ & $78.34$ \\
        & GPT-5.2~\cite{singh2025openai} & $67.67$ & $6.26$ & $8.33$ & $1.99$ & $85.73$ \\

        \midrule

        \multirow{19}{*}{MIMIC-IV-ECG}& PULSE (7B)~\cite{liu2024teach} & $72.22$ & $3.75$ & $3.75$ & $1.22$ & $29.06$ \\
        & GEM (7B)~\cite{lan2025gem} & $76.45$ & $4.35$ & $4.91$ & $1.31$ & $58.29$ \\
        & ECG-R1-SFT (8B)~\cite{jin2026ecg} & $80.02$ & $5.84$ & $5.92$ & $1.72$ & $22.21$ \\
        & ECG-R1-RL (8B)~\cite{jin2026ecg} & \scalebox{1.1}{$\bf 80.41$} & \scalebox{1.1}{$\bf 6.07$} & $6.28$ & $1.72$ & $23.01$ \\
        & OpenTSLM (3B)~\cite{langer2025opentslm} & $51.56$ & $0.39$ & $0.39$ & $0.14$ & $35.52$ \\
        \cmidrule{2-7}
        & Hulu-Med (7B)~\cite{jiang2025hulu} & $51.59$ & $4.79$ & $6.13$ & $1.16$ & $86.01$ \\
        & Hulu-Med (32B)~\cite{jiang2025hulu} & $51.56$ & $3.16$ & $4.50$ & $1.54$ & \scalebox{1.1}{$\bf 97.02$} \\
        & MedGemma (4B)~\cite{sellergren2025medgemma} & $49.57$ & $0.83$ & $0.92$ & $0.80$ & $30.37$ \\
        & MedGemma (27B)~\cite{sellergren2025medgemma} & $50.88$ & $1.28$ & $2.11$ & $1.46$ & $79.19$ \\
        & MedGemma-1.5 (4B)~\cite{sellergren2025medgemma} & $47.66$ & $1.01$ & $1.25$ & $0.78$ & $81.81$ \\
        \cmidrule{2-7}
        & Qwen3-VL (8B)~\cite{bai2025qwen3} & $53.50$ & $5.39$ & $6.07$ & $1.54$ & $86.90$ \\
        & Qwen3-VL (32B)~\cite{bai2025qwen3} & $54.24$ & $5.09$ & $6.07$ & $1.59$ & $88.06$ \\
        & Llama-3.2-Vision (11B)~\cite{grattafiori2024llama} & $49.93$ & $1.34$ & $1.70$ & $1.05$ & $75.26$ \\
        & Llama-3.2-Vision (90B)~\cite{grattafiori2024llama} & $52.01$ & $3.39$ & $4.53$ & $1.28$ & $87.82$ \\
        \cmidrule{2-7}
        & Gemini-2.5-Flash~\cite{comanici2025gemini} & $53.26$ & $4.97$ & $5.78$ & $1.78$ & $89.82$ \\
        & Gemini-2.5-Pro~\cite{comanici2025gemini} & $54.60$ & $3.39$ & $5.09$ & $1.96$ & $88.39$ \\
        & Gemini-3-Flash~\cite{team2023gemini} & $57.84$ & $5.92$ & \scalebox{1.1}{$\bf 9.68$} & \scalebox{1.1}{$\bf 2.02$} & $91.22$ \\
        & GPT-5-Mini~\cite{singh2025openai} & $52.63$ & $2.29$ & $3.07$ & $1.70$ & $75.02$ \\
        & GPT-5.2~\cite{singh2025openai} & $64.22$ & $5.63$ & $7.62$ & $1.95$ & $83.42$ \\
        \bottomrule
    \end{tabular}
    }
\end{table}

\paragraph{Evaluation metrics}
The evaluation is conducted in a multi-turn conversational format to assess the model's step-by-step reasoning process.
Rather than relying on rigid string matching, we employ Gemini-3-Flash~\cite{team2023gemini} to match each response to the predefined answer options and compare the selected answer with the ground truth.
This validator performs answer matching rather than judging the clinical quality of free-form reasoning.
We quantify the models' reasoning capabilities using the following metrics:
% \vspace{-2mm}
\begin{itemize}[leftmargin=0.5cm]
    \item \textbf{Initial Diagnosis Accuracy (IDA) (\%)}: Measures the model's baseline diagnostic accuracy based solely on the Initial Diagnostic Question, prior to engaging in any step-wise reasoning.
    \item \textbf{Completion (\%)}: Measures whether the model navigates the entire causal chain without a single logical or perceptual failure, with evaluation terminating upon the first incorrect response at any reasoning step. We recognize that this complete-traversal requirement may be overly conservative, as it demands not only perceptual or grounding capability but also diagnostic decision capability from the model. To mitigate this, we report two variants that separate these two capabilities:
    \begin{itemize}[leftmargin=0.5cm]
        \item \textbf{Completion (Hard)}: Any incorrect response at any reasoning step is treated as a failure.
        \item \textbf{Completion (Soft)}: A wrong response at a Diagnostic Decision step whose ground truth is ``further findings are required to confirm the diagnosis'' (\textit{i.e.,} a non-terminal loop) is tolerated --- the ground-truth answer is appended to the conversation history and the evaluation proceeds to the next finding loop. Otherwise, a wrong response still terminates the evaluation, as in Hard.
    \end{itemize}
    \item \textbf{Depth (0-4)}: Quantifies the model's average progression through the 4-step verification sequence, evaluated independently per finding. If a model fails at any step within a finding loop, its depth score for that finding is recorded, and the evaluation proceeds to the next required finding by providing the ground-truth history of all preceding steps in the prompt. These finding-level depths are then pooled across all samples to calculate the global average, allowing us to measure which stage the models can successfully reach on average. To provide clarity on this scoring rule, a step-by-step example is provided in Appendix~\ref{sec:app_depth_example}.
    \item \textbf{GT-Reasoning-Based Diagnosis Accuracy (GT-RDA) (\%)}: Assesses the model's diagnostic accuracy when, regardless of any prior failures, the complete ground-truth reasoning history up to the final Diagnostic Decision step is explicitly provided in the prompt. By comparing this accuracy with the Initial Diagnosis Accuracy, we measure the model's capacity to comprehend and leverage the provided reasoning process.
\end{itemize}

\subsection{Results}

\paragraph{Evaluation of reasoning completion}

Table~\ref{tab:result_main} highlights a significant limitation in the models' ability to sustain a complete chain of clinical reasoning.
Under \textbf{Completion (Hard)}, performance is remarkably poor across all evaluated models and both datasets, with the highest success rates reaching only around $7\%$.
Under \textbf{Completion (Soft)}, the highest success rates also remain limited to around $11\%$.
This implies that models lack not only diagnostic decision capability but also perceptual and grounding capability itself.
Notably, despite being explicitly tailored to the medical or ECG domains, smaller models such as OpenTSLM (3B), MedGemma (4B), and MedGemma-1.5 (4B) rarely succeed, with completion rates around or below 1\% under both variants.
A qualitative review of their responses reveals that these compact models frequently generate irrelevant or repetitive responses during the verification loops.
% This suggests that their limited parameter capacity restricts their ability to maintain sustained logical reasoning.
Interestingly, Qwen3-VL (8B) shows relatively robust performance, suggesting that its training corpus encompassed medical or ECG-specific data. By contrast, Llama-3.2-Vision (11B) underperforms despite its larger size, likely due to limited prior exposure to such data.
To contextualize these completion rates, we evaluate whether physicians can answer the same questions from the rendered ECGs and quantify the success attainable by random answering throughout the protocol.
The human and random-guessing experiments are reported in Appendices~\ref{sec:app_human_baseline} and~\ref{sec:app_random_baseline}, respectively.

\paragraph{Depth analysis}

% The \textbf{Depth} metric isolates the exact stage where the models' reasoning sequences collapse on average.
Consistently across both datasets, with the exception of the smallest models (OpenTSLM (3B), MedGemma (4B), MedGemma-1.5 (4B)), all evaluated models achieve an average depth greater than 1.0.
This indicates a broad success in the \textbf{Criterion Selection} stage, which means the models successfully identify what clinical criteria must be evaluated to make a diagnosis based on medical knowledge.
However, average depth scores rarely exceed 2.0 across the board, pinpointing the critical bottleneck at the subsequent \textbf{Finding Identification} and \textbf{ECG Grounding} stages.
This pattern suggests that the primary limitation in automated ECG interpretation is not a knowledge deficit but a substantive reasoning gap, where current models lack the ability to reliably link diagnostic criteria to the granular detection and measurement of specific findings within the signal.
To examine whether this bottleneck depends on the multiple-choice formulation, we additionally evaluate grounding without segment or measurement options (Appendix~\ref{sec:app_free_form_grounding}).
We also withhold the ECG or replace it with an unrelated recording to test dependence on the target signal (Appendix~\ref{sec:app_control}).
The protocols, results, and analyses of these supplementary experiments are presented in the respective appendices.

\paragraph{IDA vs. GT-RDA}

\begin{figure}[t]
    \centering
    \begin{subfigure}[b]{0.49\linewidth}
        \centering
        \includegraphics[width=\linewidth]{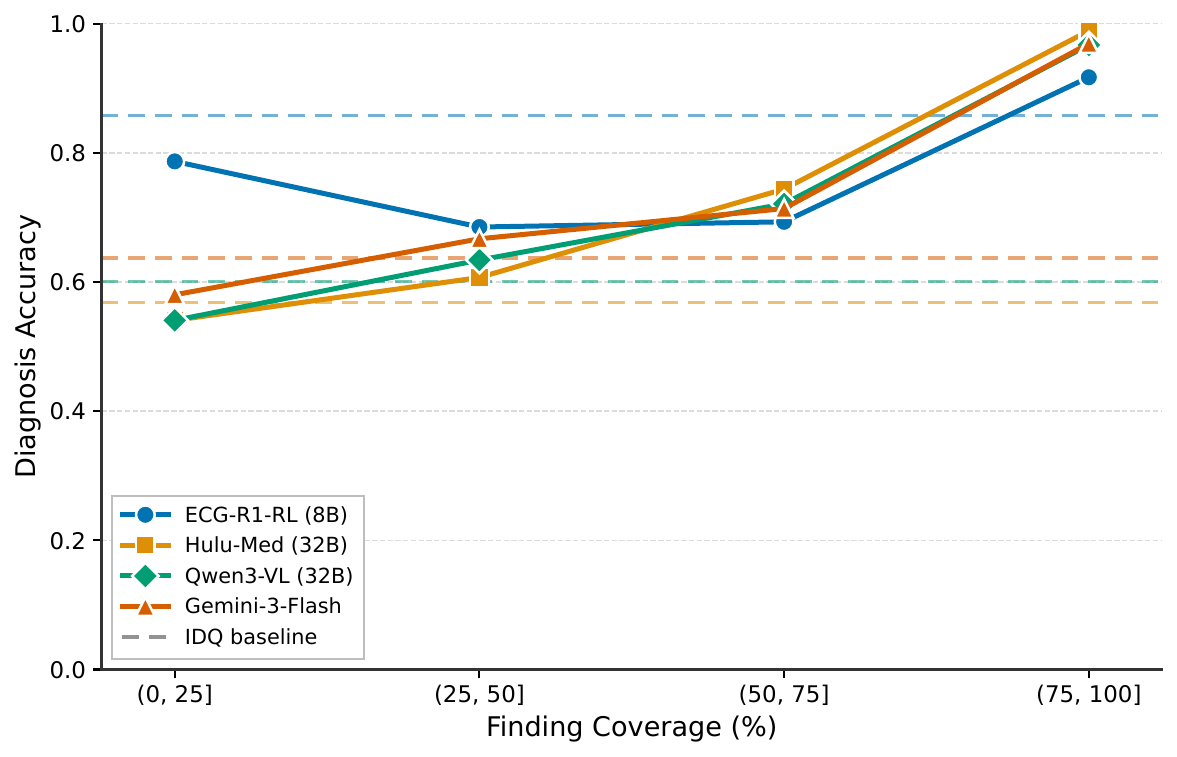}
        \caption{PTB-XL}
    \end{subfigure}
    \hfill
    \begin{subfigure}[b]{0.49\linewidth}
        \centering
        \includegraphics[width=\linewidth]{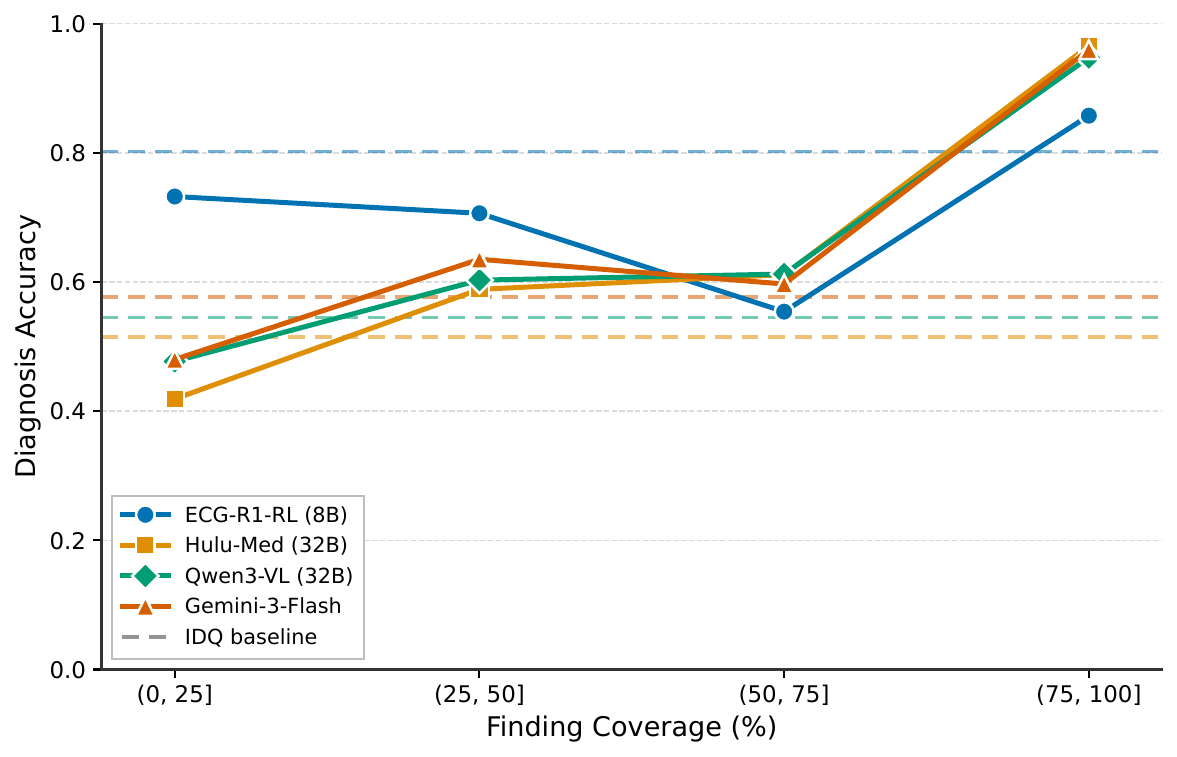}
        \caption{MIMIC-IV-ECG}
    \end{subfigure}
    \caption{Diagnosis accuracy under progressive finding coverage. Each curve shows the binary diagnosis accuracy of a representative model for each category, where the model is teacher-forced with the ground-truth reasoning history through a varying fraction of the required findings. Each point is the average accuracy within one quartile bin of finding coverage. Dashed horizontal lines indicate each model's Initial Diagnosis Accuracy (IDA) baseline.}
    \label{fig:fc}
    % \vspace{-4mm}
\end{figure}

% Another notable insight of this benchmark emerges from the comparison of Initial Diagnosis Accuracy (IDA) and GT-Reasoning-Based Diagnosis Accuracy (GT-RDA).
As a baseline observation, excluding ECG-specific models, most evaluated models achieve an IDA of approximately 50\%, indicating they are randomly guessing the initial diagnosis.
However, for these non-ECG-specific models, providing the ground-truth reasoning trajectory yields a massive surge in diagnostic accuracy across both datasets (\textit{e.g.,} Hulu-Med (32B) increasing from $57.13\%$ to $99.29\%$ on PTB-XL and $51.56\%$ to $97.02\%$ on MIMIC-IV-ECG).
This demonstrates their latent capacity to utilize valid evidence when fully guided.
In contrast, the ECG-specific models exhibit a performance collapse when provided with the correct reasoning process.
For instance, ECG-R1-RL achieves a high IDA ($85.86\%$ on PTB-XL and $80.41\%$ on MIMIC-IV-ECG), yet its accuracy drops significantly under the GT-RDA evaluation ($22.60\%$ and $23.01\%$, respectively).
% \textcolor{blue}{PULSE exhibits the same pattern across both datasets.}
Manual inspection of their responses revealed that these ECG-specific MLLMs are highly sensitive to changes in instruction format.
Their fine-tuning on narrow instruction formats appears to limit their abilit to interpret the unfamiliar ``further findings are required'' option, contributing to the observed performance degradation under GT-RDA.
The corresponding binary-decision experiment is presented in Section~\ref{sec:fc_analysis} and Figure~\ref{fig:fc}.

% This unexpected degradation reveals a notable issue in current ECG-specific MLLMs, in that they have not learned to interpret ECGs through valid clinical reasoning.
% Instead, they appear to have optimized for a superficial pattern that links global signal patterns directly to high-level diagnostic conclusions.
% While training these models to generate interpretations may yield high subjective scores under the LLM-as-a-Judge framework, our benchmark empirically demonstrates that such approaches struggle to instill the genuine, step-by-step clinical deduction required for reliable ECG interpretation.

\subsection{Diagnosis accuracy under progressive finding coverage}
\label{sec:fc_analysis}

The metric-based evaluation in Section~\ref{sec:expr_setup} reflects a key design assumption of our benchmark, that accurate diagnosis becomes attainable only after every required finding has been identified, with each constituent finding carrying substantive diagnostic information rather than serving as a redundant verification step.
To verify this assumption empirically, we conduct an additional experiment that traces how diagnostic accuracy changes as the model is provided with increasing amounts of ground-truth findings.
Specifically, for each sample with $K$ finding loops, we measure the model's diagnostic accuracy at every loop index $k \in \{1, \ldots, K\}$. At each $k$, the model is provided with the ground-truth reasoning history through the first $k$ loops and asked for a binary (Yes/No) diagnostic decision while the ``further findings are required to confirm the diagnosis'' option is excluded from the answer choices. The resulting accuracy data points are then bucketed by their finding-coverage ratio $k/K$ into four quartile bins.
Full implementation details of this protocol are provided in Appendix~\ref{sec:app_fc_protocol}.

Figure~\ref{fig:fc} reports the resulting curves for a representative model from each category across both PTB-XL and MIMIC-IV-ECG.
ECG-R1-RL exhibits a notably higher initial diagnosis accuracy (the dashed IDA baseline) than the other models, which we attribute to the fact that it has been instruction-tuned on PTB-XL and MIMIC-IV-ECG and therefore benefits from direct exposure to these datasets during training.
More importantly, when the full set of required findings is provided (the rightmost bin, $75$--$100\%$ coverage), all four models attain high diagnostic accuracy on both source datasets.
This convergence validates the design of our benchmark: the reasoning path is structured such that accurate diagnosis becomes attainable when every required finding has been identified, which in turn confirms that the constituent findings carry genuine diagnostic information rather than serving as redundant verification steps.
Another point worth noting is the stark contrast between ECG-R1-RL's rightmost-bin accuracy ($\approx 90\%$) and its low GT-RDA ($\approx 23\%$).
This gap is caused by the answer-choice restriction, as this experiment excludes the ``further findings are required to confirm the diagnosis'' option that GT-RDA exposes. Ultimately, this reveals that ECG-R1-RL frequently selects ``further findings are required to confirm the diagnosis'' at the final Diagnostic Decision, even when the complete ground-truth reasoning history is already provided.
To examine the role of response format and differences in the cases contributing to each coverage bin, we also conduct further analyses in Appendix~\ref{sec:app_fc_protocol}

\section{Discussion}
\label{sec:discussion}

This study addresses a key blind spot in the evaluation of multimodal large language models for automated ECG interpretation by introducing \textbf{ECG-Reasoning-Benchmark}, which shifts the evaluation paradigm from the LLM-as-a-Judge framework to rigorously verifying the clinical deduction process.
Our comprehensive analysis across a diverse suite of models reveals that while current state-of-the-art models excel at generating plausible diagnoses, they fail to ground their logic to granular visual evidence in the ECG signal.
Furthermore, the performance drop of ECG-specific models under GT-RDA underscores that high initial diagnostic accuracy does not guarantee reliable performance in structured verification.
These findings highlight that satisfying the linguistic evaluation of the LLM-as-a-Judge framework is insufficient to ensure clinical reliability, and future development must prioritize transparent, step-by-step reasoning grounded in the physical signal to achieve trustworthy automated ECG interpretation.
Despite these contributions, we acknowledge several limitations in its current implementation.

\paragraph{The exclusion of diagnostic uncertainty}
Although real-world ECG interpretation inherently involves diagnostic uncertainty and ambiguity, we restricted the benchmark to samples on which the dataset-provided label and our pipeline's diagnosis agree, and excluded borderline or ambiguous samples through manual review to reduce ambiguity in the reference answers.
The excluded cases include noisy recordings, borderline measurements near diagnostic thresholds (\textit{e.g.,} a PR interval of 205 ms), and complex presentations with coexisting conditions.
While this design choice provides a strict standard for assessing logical deduction, it limits the benchmark's coverage of the full clinical distribution and its ability to evaluate how models navigate uncertainty.
The benchmark therefore focuses on unambiguous, textbook-defined presentations and is intended as a test of necessary minimum competence rather than a sufficient demonstration of real-world clinical proficiency.
Future iterations aim to formalize uncertainty and reintroduce borderline cases, enabling evaluation of a model's capacity to recognize ambiguity and express appropriate diagnostic doubt.

\paragraph{Structured verification vs. clinical heuristics}
To systematically evaluate reasoning capabilities, our benchmark enforces a formalized reasoning process whose finding sequences are derived from textbook definitions and validated by board-certified internal medicine specialists.
However, this structured multi-turn evaluation may not perfectly mirror the dynamic workflow of experienced clinicians.
In practice, physicians frequently rely on clinical heuristics rather than exhaustively verifying every criterion.
For instance, a clinician may confidently confirm a diagnosis based on a highly certain finding without exhausting the remaining criteria.
Our benchmark currently requires complete traversal of the associated findings, penalizing models that attempt such heuristic shortcuts.
Although Completion (Soft) partially mitigates this gap, it does not weight findings by clinical severity or certainty.
The resulting scores should therefore be interpreted as measures of structured verification alongside conventional diagnostic accuracy.
In future work, we plan to incorporate severity and certainty weightings into our reasoning diagrams, supporting an early termination evaluation scheme that better reflects the workflow of experienced clinicians.

\begin{ack}
This work was supported by the Institute of Information \& Communications Technology Planning \& Evaluation (IITP) grants (No.RS-2019-II190075), the Korea Health Industry Development Institute (KHIDI) grant (No.RS-2025-02213750), the InnoCORE Program (N10250156), and the National Research Foundation of Korea (NRF) grant (NRF-2020H1D3A2A03100945, RS-2026-25484088), and the Medical Scientist Training Program funded by the Korean government (MSIT, MOHW).
\end{ack}

\newpage
{
\small
\bibliographystyle{plain}
\bibliography{neurips_2026}
}

%%%%%%%%%%%%%%%%%%%%%%%%%%%%%%%%%%%%%%%%%%%%%%%%%%%%%%%%%%%%

\appendix
\newpage

\section{Details of automated ECG analysis pipeline}

\subsection{Post-processing algorithms}
\label{sec:app_postproc}

The wave detection module described in Section~\ref{sec:segmentation} pairs the raw U-Net3+ output with three context-aware post-processing algorithms that together enforce physiological plausibility and exploit the multi-lead structure of the 12-lead ECG.
The full details of each component, summarized in the main text, are described below.
\paragraph{P-wave recovery via template matching}
We observed that deep-learning-based models often fail to detect non-conducted P waves that appear without a subsequent QRS complex (\textit{e.g.,} in high-degree AV blocks).
To address these missed detections, the pipeline performs a secondary search within RR intervals where no P waves were initially identified.
This targeted search utilizes SciPy's~\cite{2020SciPy-NMeth} peak detection algorithm on the unannotated segments, guided by a ``P wave template'' derived from the average duration and amplitude of successfully detected P waves within the same lead.
Specifically, candidate peaks are validated based on two criteria: (1) physiological constraints, requiring a minimal duration of 60 ms and an amplitude exceeding a noise threshold (5\% of the adjacent QRS amplitude), and (2) morphological similarity to the established template (\textit{i.e.,} sharing the identical positive, negative, or biphasic deflection).
\paragraph{Physiological constraint enforcement}
We apply strict biological rules to eliminate artifacts, such as ensuring each cardiac cycle contains only one T-wave following a QRS complex.
Multiple T wave candidates within a single RR interval are resolved by selecting the most probable peak based on its timing relative to the QT interval.
\paragraph{Multi-lead consensus alignment}
To account for lead-specific noise, we implement a 4-lead consensus rule.
A wave is validated only if it is detected at a consistent temporal location in at least 4 of the 12 leads.
Once validated, global boundaries are defined by the earliest onset and latest offset across the contributing leads to capture the full duration of the corresponding waves.

\subsection{Wave detection and segmentation performance}
\label{sec:app_seg_result}

To rigorously evaluate the wave detection module introduced in Section~\ref{sec:pipeline}, we utilized the Lobachevsky University Electrocardiography Database (LUDB)~\cite{kalyakulina2020ludb}.
The dataset was randomly partitioned into training, validation, and test sets with an 8:1:1 ratio.
To ensure the robustness of our results, this split was repeated using three different random seeds, and we report the mean performance along with the 95\% confidence intervals evaluated on the test sets.

Following the recommendation of the Association for the Advancement of Medical Instrumentation (AAMI)~\cite{aami1999ec57}, a detection was classified as a True Positive if the predicted waveform onset or offset fell within a 150 ms tolerance window of the expert ground truth annotation.
Consequently, a GT annotation lacking a corresponding prediction within this 150 ms window was counted as a False Negative, whereas a prediction without a proximate GT annotation was deemed a False Positive.

\begin{table}[h]
    \centering
    \caption{Quantitative performances for the ECG segmentation task. The evaluation was conducted on the LUDB dataset~\cite{kalyakulina2020ludb}. Mean and 95\% confidence interval are shown across 3 different train, validation, and test splits. We followed the recommendations of the Association for the Advancement of Medical Instrumentation~\cite{aami1999ec57}, where it is considered that an onset or an offset is detected correctly if their deviation from the GT annotations does not exceed the tolerance of 150 ms.}
    \label{tab:app_seg_result}
    \vspace{3mm}
    \resizebox{\textwidth}{!}{
    \begin{tabular}{cccccccc}
        \toprule
        \multirow{2}{*}{Method} & \multirow{2}{*}{Metric} & \multicolumn{2}{c}{P Wave} & \multicolumn{2}{c}{QRS Complex} & \multicolumn{2}{c}{T Wave} \\
        \cmidrule{3-8}
        & & P Onset & P Offset & QRS Onset & QRS Offset & T Onset & T Offset \\
        \midrule
        \multirow{2}{*}{ecgpuwave~\cite{goldberger2000physiobank}} & Recall & $0.872_{\pm 0.013}$ & $0.872_{\pm 0.013}$ & $0.993_{\pm 0.003}$ & $0.992_{\pm 0.003}$ & $0.890_{\pm 0.008}$ & $0.848_{\pm 0.010}$ \\
        & Precision & $0.818_{\pm 0.029}$ & $0.821_{\pm 0.030}$ & $0.998_{\pm 0.002}$ & $0.997_{\pm 0.002}$ & $0.983_{\pm 0.008}$ & $0.911_{\pm 0.001}$ \\
        \midrule
        \multirow{2}{*}{NeuroKit2~\cite{Makowski2021neurokit}} & Recall & $0.969_{\pm 0.005}$ & $0.967_{\pm 0.006}$ & $0.861_{\pm 0.011}$ & $0.796_{\pm 0.017}$ & $0.781_{\pm 0.018}$ & $0.722_{\pm 0.011}$ \\
        & Precision & $0.762_{\pm 0.040}$ & $0.762_{\pm 0.038}$ & $0.854_{\pm 0.012}$ & $0.789_{\pm 0.018}$ & $0.908_{\pm 0.018}$ & $0.836_{\pm 0.024}$ \\
        \midrule
        \multirow{2}{*}{UNet3+~\cite{joung2024deep}} & Recall & $0.972_{\pm 0.004}$ & $0.972_{\pm 0.004}$ & $0.999_{\pm 0.001}$ & $0.999_{\pm 0.001}$ & $0.983_{\pm 0.006}$ & $0.981_{\pm 0.006}$ \\
        & Precision & $\bf 0.942_{\pm 0.015}$ & $\bf 0.944_{\pm 0.015}$ & $0.999_{\pm 0.001}$ & $0.999_{\pm 0.001}$ & $\bf 0.993_{\pm 0.003}$ & $\bf 0.991_{\pm 0.003}$ \\
        \midrule
        \multirow{2}{*}{Ours} & Recall & $\bf 0.978_{\pm 0.002}$ & $\bf 0.979_{\pm 0.002}$ & $\bf 1.000_{\pm 0.000}$ & $\bf 1.000_{\pm 0.000}$ & $\bf 0.997_{\pm 0.001}$ & $\bf 0.995_{\pm 0.001}$ \\
        & Precision & $0.934_{\pm 0.036}$ & $0.940_{\pm 0.036}$ & $\bf 1.000_{\pm 0.000}$ & $\bf 1.000_{\pm 0.000}$ & $0.993_{\pm 0.005}$ & $0.991_{\pm 0.005}$ \\
        \bottomrule
    \end{tabular}
    }
\end{table}

For performance comparison, we evaluated our pipeline alongside widely used wave detection tools.
Specifically, we selected ecgpuwave~\cite{goldberger2000physiobank} and NeuroKit2~\cite{Makowski2021neurokit} as representative traditional signal processing methods.
To isolate the effect of our post-processing module, we also included the raw U-Net3+~\cite{joung2024deep} model as a deep learning baseline.
Finally, Ours denotes the complete proposed pipeline, which pairs the initial U-Net3+ outputs with our context-aware post-processing algorithms.

The quantitative results are detailed in Table~\ref{tab:app_seg_result}.
Overall, the deep learning-based approaches significantly outperformed the traditional signal processing methods.
Notably, our complete pipeline (Ours) achieved near-perfect segmentation for QRS complexes (Recall and Precision of 1.000).

When comparing Ours to the raw U-Net3+, the application of our post-processing algorithms consistently improved Recall across the waveforms.
However, we observed a slight decline in P wave Precision (\textit{e.g.,} from 0.942 to 0.934 for P Onset).
It is crucial to note that this statistical drop is an artifact of the LUDB annotation policy rather than an algorithmic failure.
Specifically, LUDB systematically omits annotations for non-conducted P waves or P waves occurring during atrioventricular (AV) dissociation.
Because our pipeline is explicitly designed to target and recover these unannotated but visually present P waves, the strict evaluation framework categorizes these correct physiological detections as False Positives, thereby artificially lowering the Precision score.

\begin{figure}[b]
    \centering
    \includegraphics[width=1.0\linewidth]{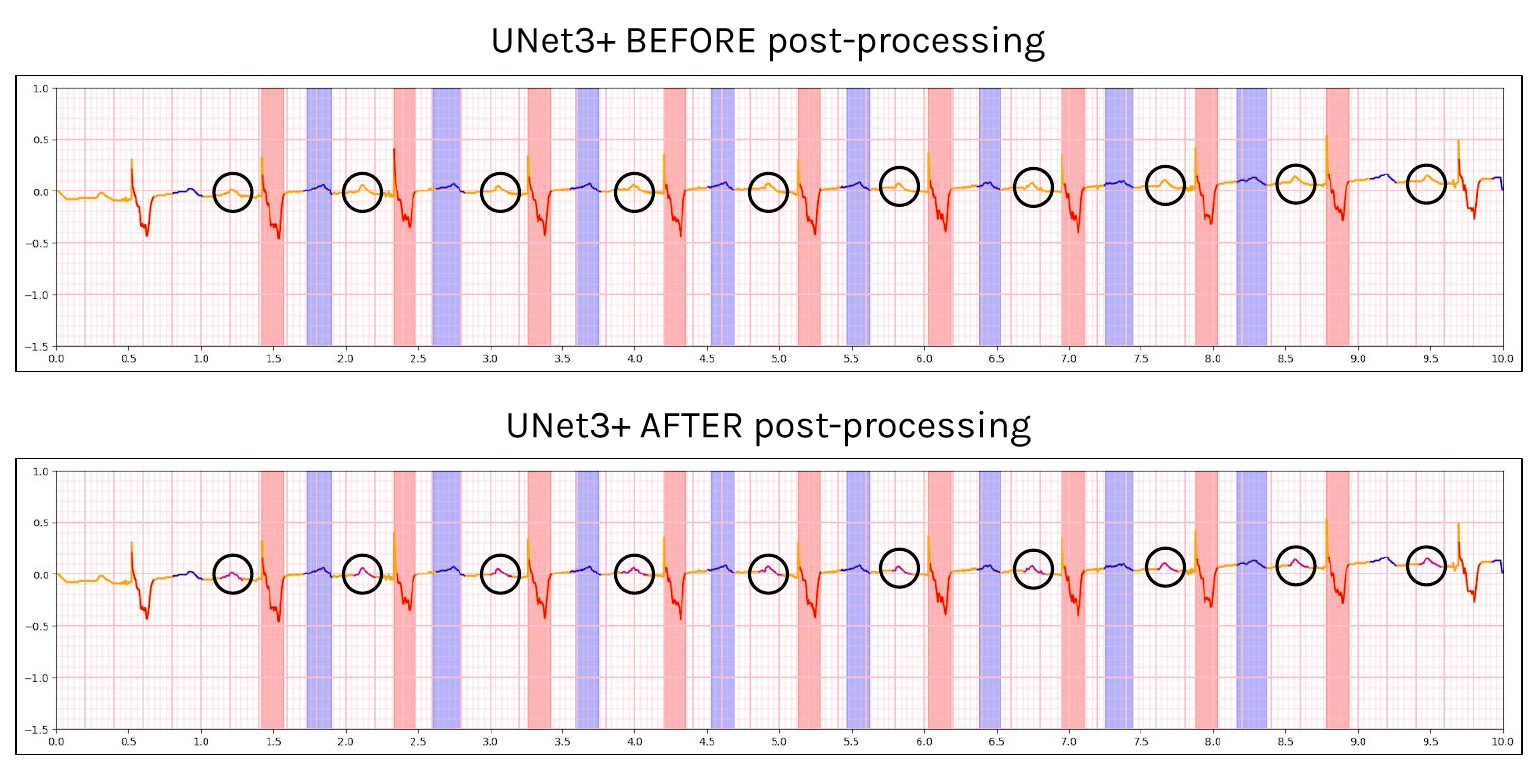}
    \caption{Comparison of P wave detection before (top) and after (bottom) post-processing. Shaded backgrounds indicate GT annotations, while colored signals show predictions. Our pipeline correctly identifies the unannotated P waves (circled), which are penalized as False Positives due to the missing GT.}
    \label{fig:app_seg_qual_result}
\end{figure}

To visually demonstrate this phenomenon, Figure~\ref{fig:app_seg_qual_result} illustrates a representative segment.
In the plots, the expert GT annotations are indicated by semi-transparent background masks (\textit{e.g.,} red for QRS complexes and blue for T waves), while the model's predictions are represented by the corresponding colored segments of the ECG signal itself.
The circled regions highlight clear, visually distinct P waves that completely lack GT annotations in the LUDB dataset (indicated by the absence of shaded masks).
As shown in the top panel, the raw U-Net3+ model strictly adheres to the learned annotation patterns and fails to detect these unannotated P waves.
In contrast, the bottom panel demonstrates that our post-processing module successfully recovers and delineates these P waves.
Because these valid detections do not have matching GT masks, they are strictly penalized as False Positives under the evaluation criteria, resulting in an artificial drop in Precision.

\subsection{Diagnostic logic diagrams for 17 core diagnoses}
\label{sec:app_diagrams}

\begin{figure}[tbp]
    \centering
    \includegraphics[width=1.0\linewidth]{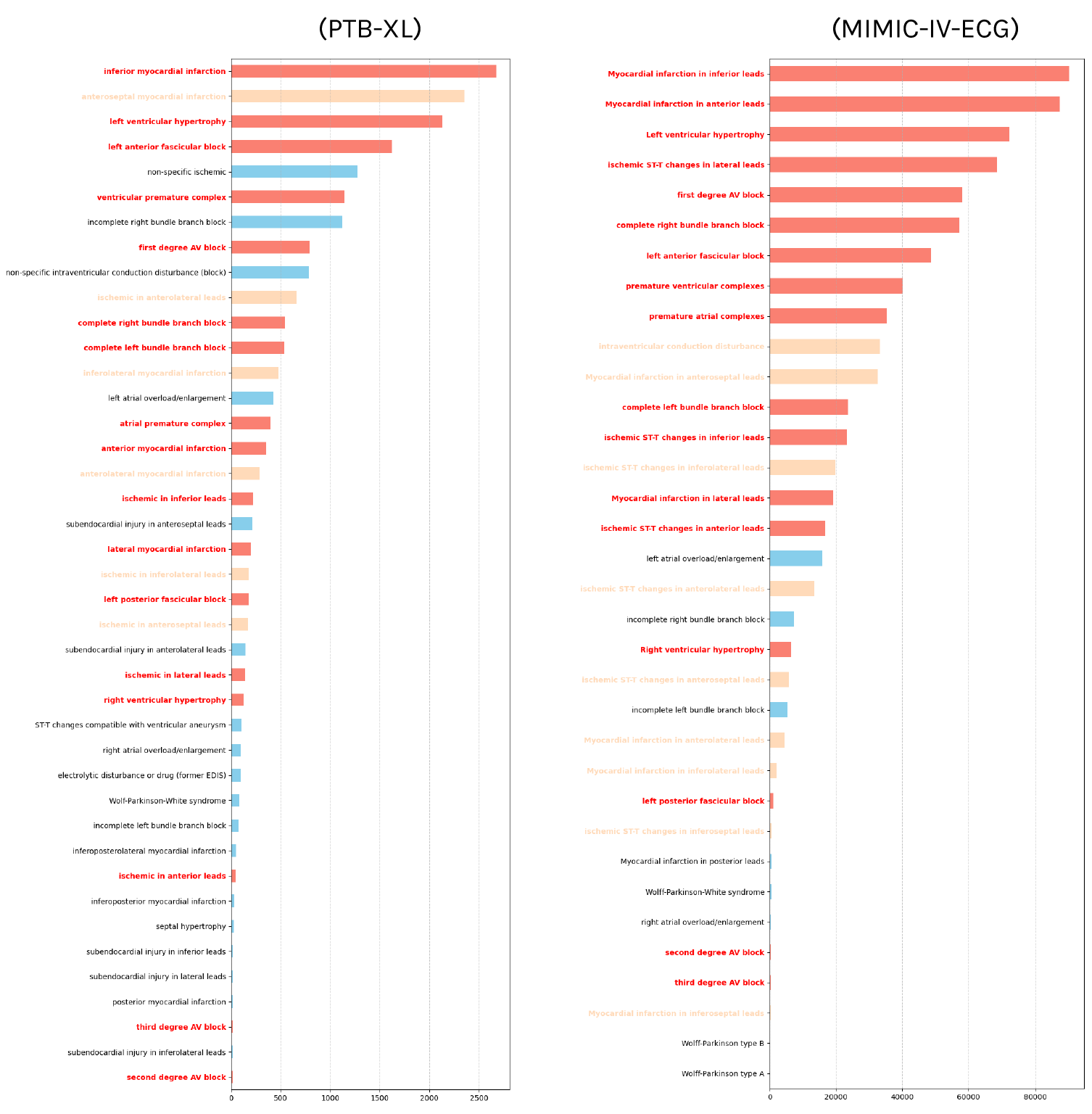}
    \caption{Prevalence of diagnostic labels in the PTB-XL (left) and MIMIC-IV-ECG (right) datasets. Diagnoses directly covered by the 17 core logic diagrams of \textbf{ECG-Reasoning-Benchmark} are highlighted in coral, while conditions that can be indirectly derived from these core diagnoses are highlighted in light coral.}
    \label{fig:histogram}
    \vspace{-2mm}
\end{figure}

The 17 core diagnoses evaluated in our benchmark comprehensively cover a wide spectrum of common and critical cardiac abnormalities.
To ensure a balanced and clinically representative evaluation, the diagnoses are systematically categorized into six major clinical groups:
\begin{itemize}
    \item \textbf{AV Block (3)}: First Degree AV Block (1AVB), Second Degree AV Block (2AVB), Third Degree AV Block (3AVB)
    \item \textbf{Conduction Disturbance (4)}: Complete Left Bundle Branch Block (CLBBB), Complete Right Bundle Branch Block (CRBBB), Left Anterior Fascicular Block (LAFB), Left Posterior Fascicular Block (LPFB)
    \item \textbf{Hypertrophy (2)}: Left Ventricular Hypertrophy (LVH), Right Ventricular Hypertrophy (RVH)
    \item \textbf{Ectopic Beat (2)}: Premature Atrial Complex (PAC), Premature Ventricular Complex (PVC)
    \item \textbf{Myocardial Infarction (3)}: Anterior Myocardial Infarction (AMI), Inferior Myocardial Infarction (IMI), Lateral Myocardial Infarction (LMI)
    \item \textbf{Ischemia (3)}: Anterior Ischemia (ISCAN), Inferior Ischemia (ISCIN), Lateral Ischemia (ISCLA)
\end{itemize}

To support the selection of these 17 core diagnoses, we analyzed their prevalence within the two source datasets.
Based on the Diagnostic SCP-codes used in the human-annotated PTB-XL dataset, Figure~\ref{fig:histogram} illustrates the distribution of diagnostic labels across both PTB-XL and MIMIC-IV-ECG.
In these histograms, the diagnoses directly evaluated by our benchmark are highlighted in coral.
We also highlight diagnoses in light coral that, while not explicitly listed among the 17 core categories, can be logically derived from them.
For instance, an ``inferolateral myocardial infarction'' can be indirectly inferred through the combination of inferior and lateral MI criteria.
Similarly, a ``(non-specific) intraventricular conduction disturbance'' can be identified when a prolonged QRS duration is present without meeting the strict morphological criteria for specific blocks like CLBBB or CRBBB.
Consequently, the 17 core diagnoses directly cover 56.05\% of the samples in PTB-XL and 82.03\% in MIMIC-IV-ECG.
When incorporating the indirectly derivable conditions, the comprehensive coverage expands significantly to 76.85\% for PTB-XL and 96.18\% for MIMIC-IV-ECG.
This high prevalence suggests that our benchmark evaluates highly clinically impactful and frequently encountered conditions in real-world settings.

The complete set of hierarchical logic diagrams for the 17 core ECG diagnoses is presented in Figure~\ref{fig:1avb}--\ref{fig:iscla}.
Note that these diagrams illustrate not only the sequence of clinical findings but also the specific grounding elements, such as wave, lead, and measurement grounding, required for each finding, reflecting the exact step-by-step evaluation structure of \textbf{ECG-Reasoning-Benchmark}.

\section{Dataset statistics and answer option construction}

\subsection{Dataset statistics}
\label{sec:app_data_stat}

\begin{table}[h]
    \centering
    \caption{Overall statistics of \textbf{ECG-Reasoning-Benchmark} derived from the PTB-XL and MIMIC-IV-ECG datasets}
    \label{tab:app_data_stat}
    \vspace{3mm}
    \begin{tabular}{lccc}
        \toprule
        \multicolumn{2}{l}{Source Dataset} & PTB-XL & MIMIC-IV-ECG \\
        \midrule
        \multicolumn{2}{l}{\# of Unique ECGs} & $2,871$ & $3,317$ \\
        \midrule
        \multirow{2}{*}{\# of Cases (Multi-turn)} & $+$ & $1,377$ & $1,655$ \\
        & $-$ & $1,707$ & $1,704$ \\
        \midrule
        \multicolumn{2}{l}{\# of Total QA pairs} & $34,145$ & $38,296$ \\
        \midrule
        \multicolumn{2}{l}{Avg. reasoning turns} & $11.07$ & $11.40$ \\
        \bottomrule
    \end{tabular}
\end{table}

Table~\ref{tab:app_data_stat} summarizes the overall characteristics of \textbf{ECG-Reasoning-Benchmark} derived from both PTB-XL and MIMIC-IV-ECG.
Specifically, the table details the total number of unique ECG recordings ($2,871$ for PTB-XL and $3,317$ for MIMIC-IV-ECG), the distribution of positive ($+$) and negative ($-$) cases evaluated through multi-turn interactions, and the vast scale of total question-answering (QA) pairs ($34,145$ and $38,296$, respectively).
Notably, the evaluation loops require an average of approximately 8 reasoning turns per case ($11.07$ for PTB-XL and $11.40$ for MIMIC-IV-ECG), highlighting the multi-step complexity of our benchmark.

To further illustrate the structural depth of our evaluation framework, Table~\ref{tab:app_data_stat_details} presents the detailed breakdown of cases for each of the 17 core diagnoses.
In the table, the specific number of cases allocated to each unique reasoning path is detailed in the ``\# Per Reasoning Path'' column, where counts are separated by a vertical pipe (|).
To ensure a balanced and robust evaluation, we set a baseline target of 100 positive ($+$) and 100 negative ($-$) cases for each diagnosis per dataset.
Crucially, we prioritized a strictly uniform distribution of cases across all valid reasoning paths to prevent models from exploiting superficial shortcuts.
Consequently, when the target of 100 was not perfectly divisible by the number of reasoning paths, the total sample size was slightly increased to maintain this exact balance (\textit{e.g.,} sampling 34 cases for each of 3 paths yields 102 total cases).
Conversely, certain rare diagnoses (\textit{e.g.,} Second and Third Degree AV Block) contain fewer cases due to their natural scarcity in the source datasets.

\begin{table}[t]
    \centering
    \caption{Detailed case distribution for the 17 core diagnoses in \textbf{ECG-Reasoning-Benchmark}. The table outlines the total number of positive ($+$) and negative ($-$) cases, the number of distinct reasoning paths for each outcome, and the specific stratification of cases across these paths for both PTB-XL and MIMIC-IV-ECG source datasets.}
    \label{tab:app_data_stat_details}
    \vspace{3mm}
    \resizebox{\textwidth}{!}{
    \begin{tabular}{cccclcl}
        \toprule
        & & \# Reasoning & \multicolumn{2}{c}{PTB-XL} & \multicolumn{2}{c}{MIMIC-IV-ECG} \\
        Dx & & Paths & \# Cases & \# Per Reasoning Path & \# Cases & \# Per Reasoning Path \\
        \midrule
        \multirow{2}{*}{1AVB} & $+$ & 1 & 100 & 100 & 100 & 100 \\
        & $-$ & 2 & 100 & 18 | 82 & 100 & 50 | 50 \\
        \midrule
        \multirow{2}{*}{2AVB} & $+$ & 1 & 3 & 3 & 49 & 49 \\
        & $-$ & 2 & 100 & 50 | 50 & 100 & 50 | 50 \\
        \midrule
        \multirow{2}{*}{3AVB} & $+$ & 2 & 9 & 0 | 9 & 100 & 50 | 50 \\
        & $-$ & 3 & 102 & 34 | 34 | 34 & 102 & 34 | 34 | 34 \\
        \midrule
        \multirow{2}{*}{CLBBB} & $+$ & 2 & 100 & 50 | 50 & 100 & 50 | 50 \\
        & $-$ & 4 & 100 & 25 | 25 | 25 | 25 & 100 & 25 | 25 | 25 | 25 \\
        \midrule
        \multirow{2}{*}{CRBBB} & $+$ & 2 & 100 & 15 | 85 & 100 & 50 | 50 \\
        & $-$ & 4 & 100 & 4 | 18 | 39 | 39 & 100 & 25 | 25 | 25 | 25 \\
        \midrule
        \multirow{2}{*}{LAFB} & $+$ & 1 & 100 & 100 & 100 & 100 \\
        & $-$ & 4 & 102 & 18 | 28 | 28 | 28 & 100 & 25 | 25 | 25 | 25 \\
        \midrule
        \multirow{2}{*}{LPFB} & $+$ & 1 & 31 & 31 & 100 & 100 \\
        & $-$ & 4 & 101 & 8 | 15 | 39 | 39 & 100 & 25 | 25 | 25 | 25 \\
        \midrule
        \multirow{2}{*}{PAC} & $+$ & 1 & 100 & 100 & 100 & 100 \\
        & $-$ & 2 & 100 & 50 | 50 & 100 & 50 | 50 \\
        \midrule
        \multirow{2}{*}{PVC} & $+$ & 1 & 100 & 100 & 100 & 100 \\
        & $-$ & 2 & 100 & 50 | 50 & 100 & 50 | 50 \\
        \midrule
        \multirow{2}{*}{LVH} & $+$ & 7 & 100 & 4 | 16 | 16 | 16 | 16 | 16 | 16 & 105 & 15 | 15 | 15 | 15 | 15 | 15 | 15 \\
        & $-$ & 4 & 100 & 25 | 25 | 25 | 25 & 100 & 25 | 25 | 25 | 25 \\
        \midrule
        \multirow{2}{*}{RVH} & $+$ & 3 & 34 & 2 | 4 | 28 & 101 & 33 | 34 | 34 \\
        & $-$ & 3 & 102 & 34 | 34 | 34 & 102 & 34 | 34 | 34 \\
        \midrule
        \multirow{2}{*}{AMI} & $+$ & 2 & 100 & 50 | 50 & 100 & 50 | 50 \\
        & $-$ & 1 & 100 & 100 & 100 & 100 \\
        \midrule
        \multirow{2}{*}{IMI} & $+$ & 2 & 100 & 50 | 50 & 100 & 50 | 50 \\
        & $-$ & 1 & 100 & 100 & 100 & 100 \\
        \midrule
        \multirow{2}{*}{LMI} & $+$ & 2 & 100 & 18 | 82 & 100 & 50 | 50 \\
        & $-$ & 1 & 100 & 100 & 100 & 100 \\
        \midrule
        \multirow{2}{*}{ISCAN} & $+$ & 2 & 100 & 43 | 57 & 100 & 50 | 50 \\
        & $-$ & 1 & 100 & 100 & 100 & 100 \\
        \midrule
        \multirow{2}{*}{ISCIN} & $+$ & 2 & 100 & 18 | 82 & 100 & 50 | 50 \\
        & $-$ & 1 & 100 & 100 & 100 & 100 \\
        \midrule
        \multirow{2}{*}{ISCLA} & $+$ & 2 & 100 & 50 | 50 & 100 & 50 | 50 \\
        & $-$ & 1 & 100 & 100 & 100 & 100 \\
        \bottomrule
    \end{tabular}
    }
\end{table}

\subsection{Answer option construction}
\label{sec:app_grounding_answer}

The Initial Diagnostic Question and Finding Identification use Yes/No answers.
Diagnostic Decision additionally offers ``further findings are required to confirm the diagnosis'' to indicate insufficient evidence for a final answer.
Grounding questions ask for affected leads, a relevant time segment, or the range containing a measured value.
We describe the construction of these grounding options because distractor selection affects how the scores should be interpreted.

\paragraph{Lead grounding}
The option set contains the leads associated with the finding and randomly selected leads to construct five options in total.
Additional leads are drawn from those that do not exhibit the target finding in the evaluated ECG.
For example, a question about ST elevation in inferior leads can present II, III, and aVF with two other leads lacking the finding.
The correct leads and distractors thus differ in the observed signal, not only in their names.

\paragraph{Wave grounding}
The target segment is accompanied by one randomly chosen segment from each of the other wave types and one segment outside the target wave.
For a QRS target, for example, the distractors include a P-wave segment, a T-wave segment, and a non-QRS interval.
For the randomly chosen segment, a duration is drawn uniformly between 160 and 300 ms and a window of that length is placed at random.
When the target is a subset of a wave type, such as a non-conducted P wave, this arbitrary segment is replaced by a segment of the same wave type that fails the target condition, such as a conducted P wave.
Wave type alone therefore cannot identify the correct answer.

\paragraph{Measurement grounding}
For each numerical feature, we examined its distribution across PTB-XL and MIMIC-IV-ECG, and defined five ranges with sufficient sample coverage.
Board-certified physicians advised on the final boundaries so that they also represent clinically interpretable values rather than arbitrary quantiles.
For prolonged QRS duration, for example, the choices are (120, 130], (130, 140], (140, 150], (150, 160], and ``above 160 ms''.
The answer is the range containing the measured target value.

\section{Implementation details}
\label{sec:app_impl_details}

\subsection{Detailed computation of the Depth metric}
\label{sec:app_depth_example}

To clarify the computation of the \textbf{Depth} metric, we present a hypothetical evaluation scenario based on the logic diagram for Complete Left Bundle Branch Block (CLBBB) described in Figure~\ref{fig:diag_clbbb}.
Specifically, suppose a positive CLBBB sample that follows the reasoning path where all four sequential findings are present (\textit{i.e.,} Yes $\rightarrow$ Yes $\rightarrow$ Yes $\rightarrow$ Yes).
According to the logic diagram, diagnosing this specific presentation requires the model to successfully navigate four distinct finding loops.

For each loop, the model undergoes the 4-step verification sequence: Criterion Selection (Step 1), Finding Identification (Step 2), ECG Grounding (Step 3), and Diagnostic Decision (Step 4).
Steps 1, 2, and 4 each contribute $1$ to the depth score upon a correct response, while Step 3 (ECG Grounding) is scored fractionally based on its $N$ required sub-tasks (\textit{i.e.,} Lead, Wave, and Measurement Grounding), where each correctly answered sub-task contributes $1/N$ to the depth score.
For example, suppose a model exhibits the following performance during the evaluation:
\begin{itemize}
    \item \textbf{Finding Loop 1 (Prolonged QRS duration)}: This finding requires $N=2$ grounding sub-tasks (Wave and Measurement Grounding). The model correctly answers Step 1 and Step 2. In Step 3, it correctly identifies the temporal segment (Wave Grounding) but selects an incorrect value range (Measurement Grounding). Upon this incorrect response, the evaluation for Loop 1 terminates.
    \begin{itemize}
        \item \textbf{Depth Score}: 1 (Step 1) + 1 (Step 2) + $1/2$ (Step 3) = 2.5
        \item Then, the exact ground-truth history for Loop 1 is explicitly injected into the prompt before proceeding to Loop 2.
    \end{itemize}
    \item \textbf{Finding Loop 2 (Dominant S waves in V1 and V2)}: This finding requires $N=1$ grounding sub-task (Wave Grounding). The model perfectly navigates this loop, passing Steps 1, 2, and 3, and successfully determining that ``further findings are required'' in Step 4.
    \begin{itemize}
        \item \textbf{Depth Score}: 1 (Step 1) + 1 (Step 2) + $1/1$ (Step 3) + 1 (Step 4) = 4.0
        \item Then, the evaluation naturally proceeds to Loop 3.
    \end{itemize}
    \item \textbf{Finding Loop 3 (Positive monophasic QRS without Q waves in lateral leads)}: This finding requires $N=2$ grounding sub-tasks (Lead and Wave Grounding). The model correctly selects the diagnostic criterion (Step 1) but incorrectly claims that the finding is absent in the current ECG (Step 2).
    \begin{itemize}
        \item \textbf{Depth Score}: 1 (Step 1) + 0 (Step 2) = 1.0
        \item Then, the true answers for Loop 3 are injected into the prompt, and the process advances to the final loop.
    \end{itemize}
    \item \textbf{Finding Loop 4 (Notched R waves in lateral leads)}: This finding requires $N=2$ grounding sub-tasks (Lead and Wave Grounding). The model passes Steps 1 and 2 but fails the initial Lead Grounding task in Step 3.
    \begin{itemize}
        \item \textbf{Depth Score}: 1 (Step 1) + 1 (Step 2) + $0/2$ (Step 3) = 2.0
        \item Then, the evaluation protocol finally terminates.
    \end{itemize}
\end{itemize}

From this single CLBBB diagnostic sample, four independent, finding-level Depth scores (2.5, 4.0, 1.0, and 2.0) are extracted.
Instead of averaging these scores to represent the sample, all four values are added directly to the global dataset pool.
For the sake of this isolated example, the final micro-averaged Depth score would be computed as $(2.5 + 4.0 + 1.0 + 2.0) / 4 = 2.375$.
In the full evaluation, this pooling strategy is applied across all samples in the dataset.
The resulting micro-average yields the final Depth metric, precisely representing the exact stage a model can successfully reach for any given clinical finding on average.

\subsection{Protocol details for the progressive finding coverage analysis}
\label{sec:app_fc_protocol}

This subsection details the protocol used for the finding coverage analysis introduced in Section~\ref{sec:fc_analysis}.
For each sample with $K$ verification loops, we construct $K$ inference instances indexed by $k \in \{1, \dots, K\}$.
For instance $k$, the model's prompt is teacher-forced with the ground-truth reasoning history of the first $k-1$ loops in full (Criterion Selection, Finding Identification, and ECG Grounding sub-tasks) plus the Criterion Selection, Finding Identification, and ECG Grounding sub-tasks of the $k$-th loop.
The model is then asked the Diagnostic Decision question of the $k$-th loop with its options restricted to \texttt{["Yes", "No"]}, which forces a binary commit and removes the ``further findings are required to confirm the diagnosis'' escape hatch otherwise available in the default protocol.
The Initial Diagnostic Question (IDQ) is run identically to the default protocol and is reported as the $0\%$-coverage baseline (drawn as a dashed horizontal line in Figure~\ref{fig:fc}).

Each binary response is then validated against the sample-level diagnosis label (\texttt{yes} or \texttt{no}) using the same Gemini-3-Flash semantic-consistency validator employed in Section~\ref{sec:expr_setup}.
Per-loop accuracy data points are bucketed by their finding-coverage ratio $(k/K) \times 100\%$ into four quartile bins ($(0, 25]$, $(25, 50]$, $(50, 75]$, $(75, 100]$), and the per-bin accuracy is computed by pooling all loops across all samples that fall into the corresponding bin.
By definition the rightmost bin $(75, 100]$ always contains every sample's final loop and therefore reflects diagnostic accuracy after the model has been provided with the entire required finding set.

\begin{table}[htbp]
\centering
\caption{IDA restricted to cases with $K\geq4$ and accuracy in the first coverage bin (\%).}
\label{tab:coverage_matched_ida}
\small
\begin{tabular}{lrrrr}
\toprule
& \multicolumn{2}{c}{PTB-XL} & \multicolumn{2}{c}{MIMIC-IV-ECG}\\
Model & IDA$_{K\geq4}$ & $(0,25]$ & IDA$_{K\geq4}$ & $(0,25]$\\
\midrule
ECG-R1-RL & 83.96 & 78.68 & 75.08 & 73.24\\
Hulu-Med (32B) & 53.63 & 54.07 & 40.83 & 41.90\\
Qwen3-VL (32B) & 53.19 & 54.07 & 46.64 & 47.71\\
Gemini-3-Flash & 58.02 & 58.02 & 42.35 & 48.01\\
\bottomrule
\end{tabular}
\end{table}

\paragraph{Coverage-bin composition}
The bins contain different case subsets because the number of required loops, $K$, varies. In particular, the first finding falls in the $(0,25]$ bin only when $K\geq4$, whereas the dashed IDA baseline averages over all cases. Recomputing IDA on cases with $K\geq4$ brings it close to first-bin accuracy for most representative models (Table~\ref{tab:coverage_matched_ida}). The lower first-bin performance therefore partly reflects case composition rather than a matched within-case decrease.

\begin{table}[htbp]
\centering
\caption{ECG-R1-RL responses giving "further findings are required" despite a Yes/No restriction (\%).}
\label{tab:coverage_out_of_format}
\small
\begin{tabular}{lrr}
\toprule
Coverage bin & PTB-XL & MIMIC-IV-ECG\\
\midrule
$(0,25]$ & 0.0 & 0.0\\
$(25,50]$ & 0.0 & 0.0\\
$(50,75]$ & 12.3 & 18.3\\
$(75,100]$ & 5.6 & 8.5\\
\bottomrule
\end{tabular}
\end{table}

\paragraph{ECG-R1-RL response format}
Each Diagnostic Decision in the progressive-coverage experiment asks for Yes or No. ECG-R1-RL nevertheless sometimes answers "further findings are required", an option used in the standard multi-turn protocol but absent from these binary questions. Such responses are scored incorrect. Their frequency peaks at 12.3\% on PTB-XL and 18.3\% on MIMIC-IV-ECG in the $(50,75]$ bin, before falling in the highest bin (Table~\ref{tab:coverage_out_of_format}). These errors contribute to the depressed middle-bin accuracy, although these rates alone do not quantify the entire effect on the curve.

The binary coverage experiment and GT-RDA also differ in their available answers: GT-RDA includes "further findings are required". Their score difference should therefore be interpreted with this format difference.
The out-of-format responses identify an instruction-following issue, while a controlled prompt ablation would be required to attribute the full GT-RDA gap to any single wording or option.

\subsection{Evaluated models and access endpoints}

To ensure the reproducibility of our benchmark results, we detail the specific model versions and access endpoints utilized in our experiments. Proprietary models were queried via their respective official APIs, utilizing the following exact model identifiers to prevent version discrepancies:
\begin{itemize}
    \item \textbf{Gemini-2.5-Flash~\cite{comanici2025gemini}}: \texttt{gemini-2.5-flash}
    \item \textbf{Gemini-2.5-Pro~\cite{comanici2025gemini}}: \texttt{gemini-2.5-pro}
    \item \textbf{Gemini-3-Flash~\cite{team2023gemini}}: \texttt{gemini-3-flash-preview}
    \item \textbf{GPT-5-Mini~\cite{singh2025openai}}: \texttt{gpt-5-mini-2025-08-07}
    \item \textbf{GPT-5.2~\cite{singh2025openai}}: \texttt{gpt-5.2-2025-12-11}
\end{itemize}

For open-weight models, we loaded the officially released model weights directly from the Hugging Face Hub.
The specific repository paths are listed below:
\begin{itemize}
    \item \textbf{PULSE (7B)~\cite{liu2024teach}}
    \begin{itemize}
        \item A multimodal LLM fine-tuned on the ECGInstruct dataset~\cite{liu2024teach}, representing the instruction-tuned model for ECGs.
        \item \textbf{Hugging Face Repository}: \texttt{PULSE-ECG/PULSE-7B}
        \item \textbf{Compute resources}: With a single RTX A6000 GPU, it resulted in a test time of 12 hours for each source dataset.
    \end{itemize}
    \item \textbf{GEM (7B)~\cite{lan2025gem}}
    \begin{itemize}
        \item A model designed for grounded ECG understanding, trained to link visual features with textual diagnoses.
        \item \textbf{Hugging Face Repository}: \texttt{LANSG/GEM}
        \item \textbf{Compute resources}: With a single RTX A6000 GPU, it resulted in a test time of 12 hours for each source dataset.
    \end{itemize}
    \item \textbf{ECG-R1 (8B)~\cite{jin2026ecg}}
    \begin{itemize}
        \item A model trained on the ECG-Protocol-Guided-Grounding-CoT dataset~\cite{jin2026ecg}, employing Reinforcement Learning to enforce a 6-step analysis protocol for structured ECG interpretation.
        \item \textbf{Hugging Face Repository}: \texttt{PKUDigitalHealth/ECG-R1-8B-SFT}, \\
        \texttt{PKUDigitalHealth/ECG-R1-8B-RL}
        \item \textbf{Compute resources}: With a single RTX A6000 GPU, it resulted in a test time of 12 hours for each source dataset.
    \end{itemize}
    \item \textbf{OpenTSLM (3B)~\cite{langer2025opentslm}}
    \begin{itemize}
        \item A time-series native language model capable of processing continuous signals directly, trained with Chain-of-Thought rationales.
        \item \textbf{Hugging Face Repository}: \texttt{OpenTSLM/llama3b-ecg-flamingo}
        \item \textbf{Compute resources}: With a single RTX A6000 GPU, it resulted in a test time of 76 hours and 84 hours, respectively for each source dataset (\textit{i.e.,} PTB-XL and MIMIC-IV-ECG).
    \end{itemize}
\end{itemize}

\begin{figure}[t]
    \centering
    \includegraphics[width=1.0\linewidth]{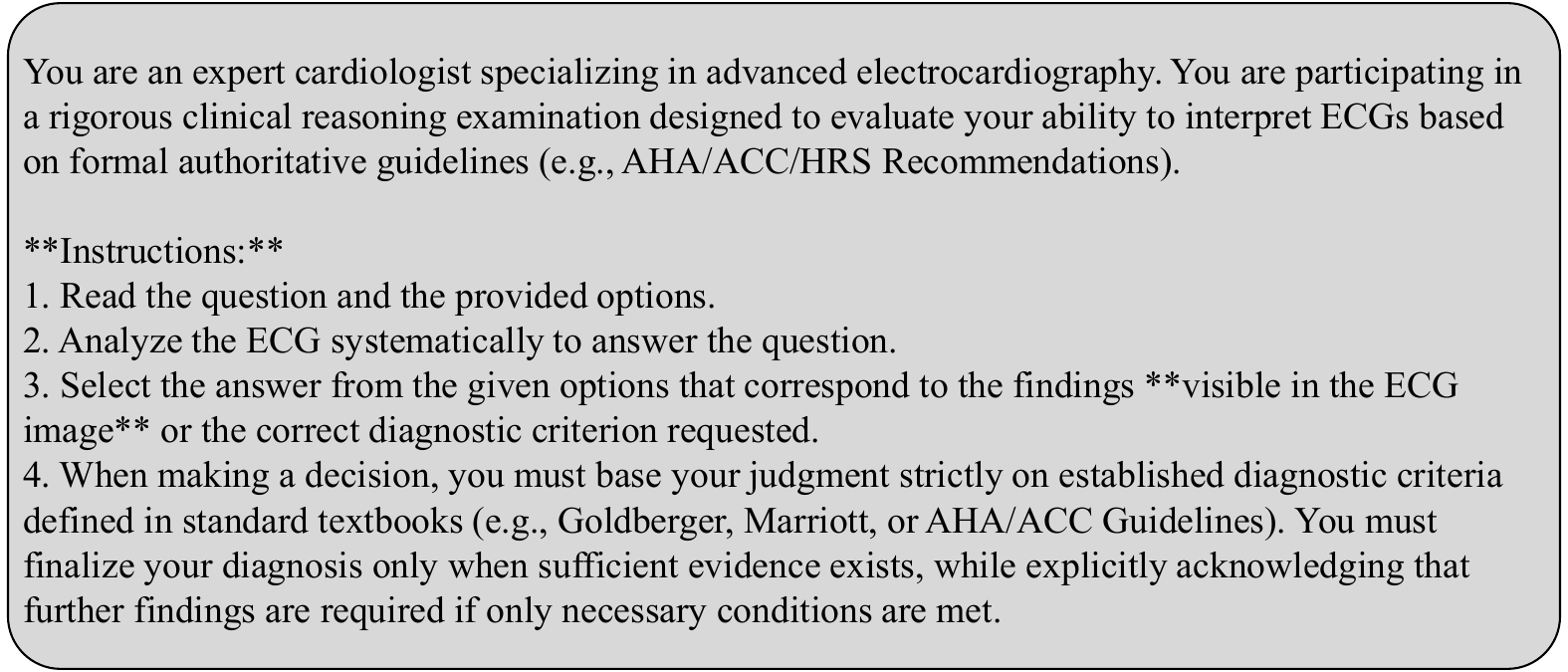}
    \caption{System prompt provided to the evaluated models}
    \label{fig:prompt}
\end{figure}

\begin{itemize}
    \item \textbf{Hulu-Med (7B, 32B)~\cite{jiang2025hulu}}
    \begin{itemize}
        \item A series of models optimized for broad biomedical tasks across various modalities, including medical text, 2D/3D images, and videos.
        \item \textbf{Hugging Face Repository}: \texttt{ZJU-AI4H/Hulu-Med-7B}, \texttt{ZJU-AI4H/Hulu-Med-32B}
        \item \textbf{Compute resources}: For the 7B-sized model (\textbf{Hulu-Med (7B)}), it resulted in a test time of 15 hours and 17 hours with a single A100 GPU, respectively for each source dataset. For the 32B-sized model (\textbf{Hulu-Med (32B)}), it resulted in a test time of 45 hours and 50 hours, respectively.
    \end{itemize}
    \item \textbf{MedGemma (4B, 27B)~\cite{sellergren2025medgemma}}, \textbf{MedGemma-1.5 (4B)~\cite{sellergren2025medgemma}}
    \begin{itemize}
        \item Instruction-tuned version of MedGemma, which is a Gemma3~\cite{team2025gemma} variant optimized for medical text and image comprehension.
        \item \textbf{Hugging Face Repository}: \texttt{google/medgemma-4b-it}, \\
        \texttt{google/medgemma-27b-it}, \texttt{google/medgemma-1.5-4b-it}
        \item \textbf{Compute resources}: For \textbf{MedGemma (4B)}, it resulted in a test time of 13 hours and 15 hours with a single RTX A6000 GPU, respectively for each source dataset. For \textbf{MedGemma (27B)}, it resulted in a test time of 44 hours and 48 hours with a single A100 GPU, respectively. For \textbf{MedGemma-1.5 (4B)}, it resulted in a test time of 110 hours and 122 hours with a single A6000 GPU, respectively.
    \end{itemize}
    \item \textbf{Qwen3-VL (8B, 32B)~\cite{bai2025qwen3}}
    \begin{itemize}
        \item The latest iteration of the Qwen-VL series, known for strong visual reasoning performance.
        \item \textbf{Hugging Face Repository}: \texttt{Qwen/Qwen3-VL-8B-Instruct}, \\
        \texttt{Qwen/Qwen3-VL-32B-Instruct}
        \item \textbf{Compute resources}: For \textbf{Qwen3-VL (8B)}, it resulted in a test time of 13 hours and 15 hours with a single A6000 GPU, respectively for each source dataset. For \textbf{Qwen3-VL (32B)}, it resulted in a test time of 20 hours and 23 hours with a single A100 GPU, respectively.
    \end{itemize}
    \item \textbf{Llama-3.2-Vision (11B, 90B)~\cite{grattafiori2024llama}}
    \begin{itemize}
        \item Meta's open-weights multimodal models, providing a robust baseline for general visual understanding.
        \item \textbf{Hugging Face Repository}: \texttt{meta-llama/Llama-3.2-11B-Vision-Instruct}, \\
        \texttt{meta-llama/Llama-3.2-90B-Vision-Instruct}
        \item \textbf{Compute resources}: For \textbf{Llama-3.2-Vision (11B)}, it resulted in a test time of 10 hours and 11 hours with a single A6000 GPU, respectively for each source dataset. For \textbf{Llama-3.2-Vision (90B)}, it resulted in a test time of 18 hours and 20 hours with three A100 GPUs, respectively.
    \end{itemize}
\end{itemize}

Additionally, \textbf{ECG-Reasoning-Benchmark} evaluates the strict logical deduction capabilities of the models rather than their creative text generation.
To maintain absolute consistency and eliminate randomness in the evaluation loop, we configured all models for deterministic output.
Specifically, the decoding temperature was set to 0 across all generation pipelines.
This configuration ensures that each model produces its most probable reasoning path consistently given the ECG context.

\subsection{System prompt design}
\label{sec:prompt}

To purely assess the models' inherent clinical reasoning and minimize variations caused by prompt engineering, we employed a standardized system prompt across all evaluated models.
Crucially, this system prompt is deliberately designed to constrain the models to follow a rigorous, textbook-defined reasoning process.
By explicitly mandating adherence to established diagnostic criteria (\textit{e.g.,} AHA/ACC guidelines and standard cardiology textbooks) and requiring sufficient visual evidence before finalizing a diagnosis, we ensure that the models are evaluated on their formal deductive logic rather than heuristic guessing or ungrounded pattern matching.
The exact prompt used to initialize the expert cardiologist persona in our evaluation loop is provided in Figure~\ref{fig:prompt}.

\section{Additional baselines and grounding analyses}

\subsection{Human baseline}
\label{sec:app_human_baseline}

To assess whether the questions can be answered from the rendered ECGs and the distractors can be distinguished, two physicians completed the same multi-turn protocol as the evaluated models.
We sampled one case per unique reasoning path: 73 from PTB-XL and 74 from MIMIC-IV-ECG.
Physician A is a general practitioner and Physician B is a board-certified internal medicine specialist.
Neither physician had access to pipeline outputs or ground-truth reasoning paths.
We evaluated GEM, ECG-R1-RL, and Gemini-3-Flash on the identical cases for a matched comparison (Table~\ref{tab:human_baseline}).

\begin{table}[t]
\centering
\caption{Human baseline and model results on the same path-stratified subsets.}
\label{tab:human_baseline}
\small
\begin{tabular}{ccccccc}
\toprule
\multirow{3}{*}{\textbf{Source Dataset}} & \multirow{3}{*}{\textbf{Model}} & \multirow{3}{*}{\textbf{IDA (\%)}} & \multicolumn{2}{c}{\textbf{Completion (\%)}} & \multirow{3}{*}{\textbf{Depth (0-4)}} & \multirow{3}{*}{\textbf{GT-RDA (\%)}} \\
\cmidrule(lr){4-5}
& & & \textbf{Hard} & \textbf{Soft} & & \\
\midrule
\multirow{5}{*}{PTB-XL}
& GEM & 79.45 & 2.74 & 2.74 & 1.15 & 53.42\\
& ECG-R1-RL & 84.93 & 8.22 & 8.22 & 1.68 & 21.92\\
& Gemini-3-Flash & 67.12 & 6.85 & 8.22 & 1.86 & 94.52\\
& Physician A & 86.30 & 56.16 & 58.90 & 3.47 & 72.60\\
& Physician B & 86.30 & 71.23 & 76.71 & 3.73 & 89.04\\
\midrule
\multirow{5}{*}{MIMIC-IV-ECG}
& GEM & 75.68 & 4.05 & 5.41 & 1.26 & 48.65\\
& ECG-R1-RL & 74.32 & 6.76 & 6.76 & 1.72 & 22.97\\
& Gemini-3-Flash & 60.81 & 6.76 & 9.46 & 1.87 & 95.95\\
& Physician A & 81.08 & 51.35 & 58.11 & 3.43 & 89.19\\
& Physician B & 81.08 & 75.68 & 77.03 & 3.81 & 93.24\\
\bottomrule
\end{tabular}
\end{table}

\begin{table}[t]
\centering
\caption{Per-question-type accuracy on the human-baseline subsets (\%).}
\label{tab:human_baseline_per_question_type}
\small
\resizebox{\linewidth}{!}{%
\begin{tabular}{cccccccc}
\toprule
\multirow{2}{*}{\textbf{Source Dataset}} & \multirow{2}{*}{\textbf{Model}} & \makecell{\textbf{Criterion} \\ \textbf{Selection}} & \makecell{\textbf{Finding} \\ \textbf{Identification}} & \makecell{\textbf{Lead} \\ \textbf{Grounding}} & \makecell{\textbf{Wave} \\ \textbf{Grounding}} & \makecell{\textbf{Measurement} \\ \textbf{Grounding}} & \makecell{\textbf{Diagnostic} \\ \textbf{Decision}}
\\
\midrule
\multirow{5}{*}{PTB-XL}
& GEM & 60.4 & 63.7 & 48.8 & 27.5 & 26.2 & 33.0\\
& ECG-R1-RL & 85.2 & 73.1 & 53.7 & 25.7 & 29.9 & 35.7\\
& Gemini-3-Flash & 96.2 & 61.0 & 68.3 & 53.3 & 36.4 & 67.0\\
& Physician A & 100.0 & 89.6 & 97.6 & 100.0 & 95.3 & 86.8\\
& Physician B & 100.0 & 93.4 & 100.0 & 99.4 & 100.0 & 92.3\\
\midrule
\multirow{5}{*}{MIMIC-IV-ECG}
& GEM & 62.7 & 64.9 & 43.9 & 26.9 & 22.2 & 30.8\\
& ECG-R1-RL & 85.4 & 69.2 & 46.3 & 37.7 & 24.1 & 36.2\\
& Gemini-3-Flash & 98.4 & 59.5 & 68.3 & 50.9 & 54.6 & 68.1\\
& Physician A & 98.4 & 87.0 & 92.7 & 99.4 & 92.6 & 88.1\\
& Physician B & 100.0 & 98.4 & 97.6 & 96.4 & 97.2 & 95.7\\
\bottomrule
\end{tabular}}
\end{table}

On PTB-XL, the physicians reached 56.16\% and 71.23\% hard completion, compared with 2.74\%--8.22\% for the three models.
On MIMIC-IV-ECG, their respective rates were 51.35\% and 75.68\%, compared with 4.05\%--6.76\%.
The higher specialist scores are consistent with sensitivity to clinical expertise.
Hard completion requires every graded step in a chain to be correct, so occasional errors accumulate across roughly eight reasoning turns per case on average.
We therefore also report accuracy by question type (Table~\ref{tab:human_baseline_per_question_type}).
Both physicians perform well on the specific grounding tasks; for example, Physician B attains 96.4\% Wave and 97.2\% Measurement Grounding accuracy on MIMIC-IV-ECG, compared with 50.9\% and 54.6\% for Gemini-3-Flash.

\subsection{Random guessing baseline}
\label{sec:app_random_baseline}

Multiple-choice questions permit occasional correct responses by change.
To quantify this floor, we evaluated a policy that chooses randomly at each question in the protocol.
Despite an IDA close to 50\%, this policy reaches only 0.24\% and 0.23\% hard completion on PTB-XL and MIMIC-IV-ECG, respectively (Table~\ref{tab:random_baseline}).
Completion is much lower than single-question accuracy because an entire reasoning path must be traversed correctly.
The model results in Table~\ref{tab:result_main} exceed this floor, although their absolute completion rates remain low.

\begin{table}[htbp]
\centering
\caption{Random-answering baseline.}
\label{tab:random_baseline}
\small
\begin{tabular}{cccccc}
\toprule
\multirow{3}{*}{\textbf{Source Dataset}} & \multirow{3}{*}{\textbf{IDA (\%)}} & \multicolumn{2}{c}{\textbf{Completion (\%)}} & \multirow{3}{*}{\textbf{Depth (0-4)}} & \multirow{3}{*}{\textbf{GT-RDA (\%)}} \\
\cmidrule(lr){3-4}
& &  \textbf{Hard} & \textbf{Soft} & & \\
\midrule
PTB-XL & 49.65 & 0.24 & 0.24 & 0.32 & 33.70\\
MIMIC-IV-ECG & 49.67 & 0.23 & 0.23 & 0.32 & 32.63\\
\bottomrule
\end{tabular}
\end{table}

\subsection{Free-form grounding}
\label{sec:app_free_form_grounding}

To examine dependence on answer options, we conducted a free-form grounding evaluation in which models directly produced segment locations and measurement values.
Gemini-3-Flash parsed the time interval or numerical value from generated response; a fixed rule then determined correctness.
For Wave Grounding, a predicted interval was correct if its intersection over union with any valid ground-truth interval was at least 0.5.
For Measurement Grounding, the parsed value had to fall within the valid grounding-truth range.
IDA and GT-RDA contain no grounding items and are unchanged by this intervention.

\begin{table}[htbp]
\centering
\caption{Free-form grounding results.}
\label{tab:free_form_grounding}
\small
\begin{tabular}{ccccccc}
\toprule
\multirow{3}{*}{\textbf{Source Dataset}} & \multirow{3}{*}{\textbf{Model}} & \multirow{3}{*}{\textbf{IDA (\%)}} & \multicolumn{2}{c}{\textbf{Completion (\%)}} & \multirow{3}{*}{\textbf{Depth (0-4)}} & \multirow{3}{*}{\textbf{GT-RDA (\%)}} \\
\cmidrule(lr){4-5}
& & & \textbf{Hard} & \textbf{Soft} & & \\
\midrule
\multirow{3}{*}{PTB-XL}
& GEM & 84.40 & 4.22 & 4.22 & 1.24 & 67.28\\
& ECG-R1-RL & 85.86 & 5.84 & 5.84 & 1.75 & 22.60\\
& Gemini-3-Flash & 63.72 & 6.32 & 7.43 & 2.01 & 91.96\\
\midrule
\multirow{3}{*}{MIMIC-IV-ECG}
& GEM & 76.45 & 4.02 & 4.49 & 1.21 & 58.29\\
& ECG-R1-RL & 80.41 & 5.15 & 5.15 & 1.66 & 23.01\\
& Gemini-3-Flash & 57.84 & 5.57 & 7.35 & 1.98 & 91.22\\
\bottomrule
\end{tabular}
\end{table}

Completion and depth decline modestly relative to the multiple-choice setting, while grounding remains the principal failure stage (Table~\ref{tab:free_form_grounding}).
This indicates that the observed bottleneck is not confined to choosing among the provided segment and measurement options.

\subsection{Visual grounding controls}
\label{sec:app_control}

We tested two controls for reliance on the target ECG.
The text-only condition withheld the ECG input.
The mismatched condition substituted an unrelated ECG sampled from a case with a different diagnosis.
The question sequence and scoring protocol were retained.
Tables~\ref{tab:app_text_only} and~\ref{tab:app_mismatched} report GEM, ECG-R1-RL, and Gemini-3-Flash on each dataset under both conditions.

\begin{table}[htbp]
\centering
\caption{Text-only control.}
\label{tab:app_text_only}
\small
\begin{tabular}{ccccccc}
\toprule
\multirow{3}{*}{\textbf{Source Dataset}} & \multirow{3}{*}{\textbf{Model}} & \multirow{3}{*}{\textbf{IDA (\%)}} & \multicolumn{2}{c}{\textbf{Completion (\%)}} & \multirow{3}{*}{\textbf{Depth (0-4)}} & \multirow{3}{*}{\textbf{GT-RDA (\%)}} \\
\cmidrule(lr){4-5}
& & & \textbf{Hard} & \textbf{Soft} & & \\
\midrule
\multirow{3}{*}{PTB-XL}
& GEM & 55.35 & 2.17 & 2.17 & 1.03 & 30.51\\
& ECG-R1-RL & 50.06 & 0.39 & 0.39 & 1.51 & 15.82\\
& Gemini-3-Flash & 55.12 & 2.50 & 5.84 & 1.88 & 93.29\\
\midrule
\multirow{3}{*}{MIMIC-IV-ECG}
& GEM & 50.73 & 2.08 & 2.08 & 0.99 & 31.20\\
& ECG-R1-RL & 50.77 & 0.83 & 0.83 & 1.53 & 15.90\\
& Gemini-3-Flash & 50.70 & 2.23 & 5.27 & 1.83 & 91.72\\
\bottomrule
\end{tabular}
\end{table}

\begin{table}[htbp]
\centering
\caption{Mismatched-ECG control.}
\label{tab:app_mismatched}
\small
\begin{tabular}{ccccccc}
\toprule
\multirow{3}{*}{\textbf{Source Dataset}} & \multirow{3}{*}{\textbf{Model}} & \multirow{3}{*}{\textbf{IDA (\%)}} & \multicolumn{2}{c}{\textbf{Completion (\%)}} & \multirow{3}{*}{\textbf{Depth (0-4)}} & \multirow{3}{*}{\textbf{GT-RDA (\%)}} \\
\cmidrule(lr){4-5}
& & & \textbf{Hard} & \textbf{Soft} & & \\
\midrule
\multirow{3}{*}{PTB-XL}
& GEM & 54.90 & 1.82 & 1.85 & 0.99 & 31.26\\
& ECG-R1-RL & 54.15 & 3.44 & 3.44 & 1.51 & 17.06\\
& Gemini-3-Flash & 55.54 & 5.35 & 9.21 & 1.85 & 91.60\\
\midrule
\multirow{3}{*}{MIMIC-IV-ECG}
& GEM & 50.22 & 1.55 & 1.55 & 0.95 & 32.18\\
& ECG-R1-RL & 51.35 & 2.80 & 2.80 & 1.44 & 17.83\\
& Gemini-3-Flash & 51.15 & 4.88 & 8.54 & 1.78 & 90.86\\
\bottomrule
\end{tabular}
\end{table}

The results reveal a clear dissociation between global perception and fine-grained grounding.
In both control settings, the Initial Diagnosis Accuracy (IDA) collapses to near random chance ($\sim$50\%), in stack contrast to the 63.72\% to 85.86\% achieved when the correct images are provided.
Completion rates also fall sharply; for instance, ECG-R1-RL's completion on PTB-XL drops from 6.42\% to 0.39\% under the text-only condition.
This confirms that the models do read the ECG and rely heavily on it for their initial, high-level diagnosis.

Crucially, however, the Depth metric exhibits a distinctly contrasting trend. Across both control conditions and datasets, Depth retains 72\% to 91\% of the baseline score achieved with the correct image.
This indicates that the model's progression through the reasoning chain halts at nearly the same stage regardless of whether a valid ECG is present.
We interpret this as evidence that current models utilize the ECG merely for a global diagnostic impression.
Once the protocol demands precise localization, such as identifying a specific wave segment or extracting a measurement, the presence of the correct image provides little benefit.
This supports the core conclusion of our paper: the models' primary failure lies specifically in fine-grained visual grounding rather than in the general perception of the image as a whole.

\begin{figure}[bp]
    \centering
    \includegraphics[width=1.0\linewidth]{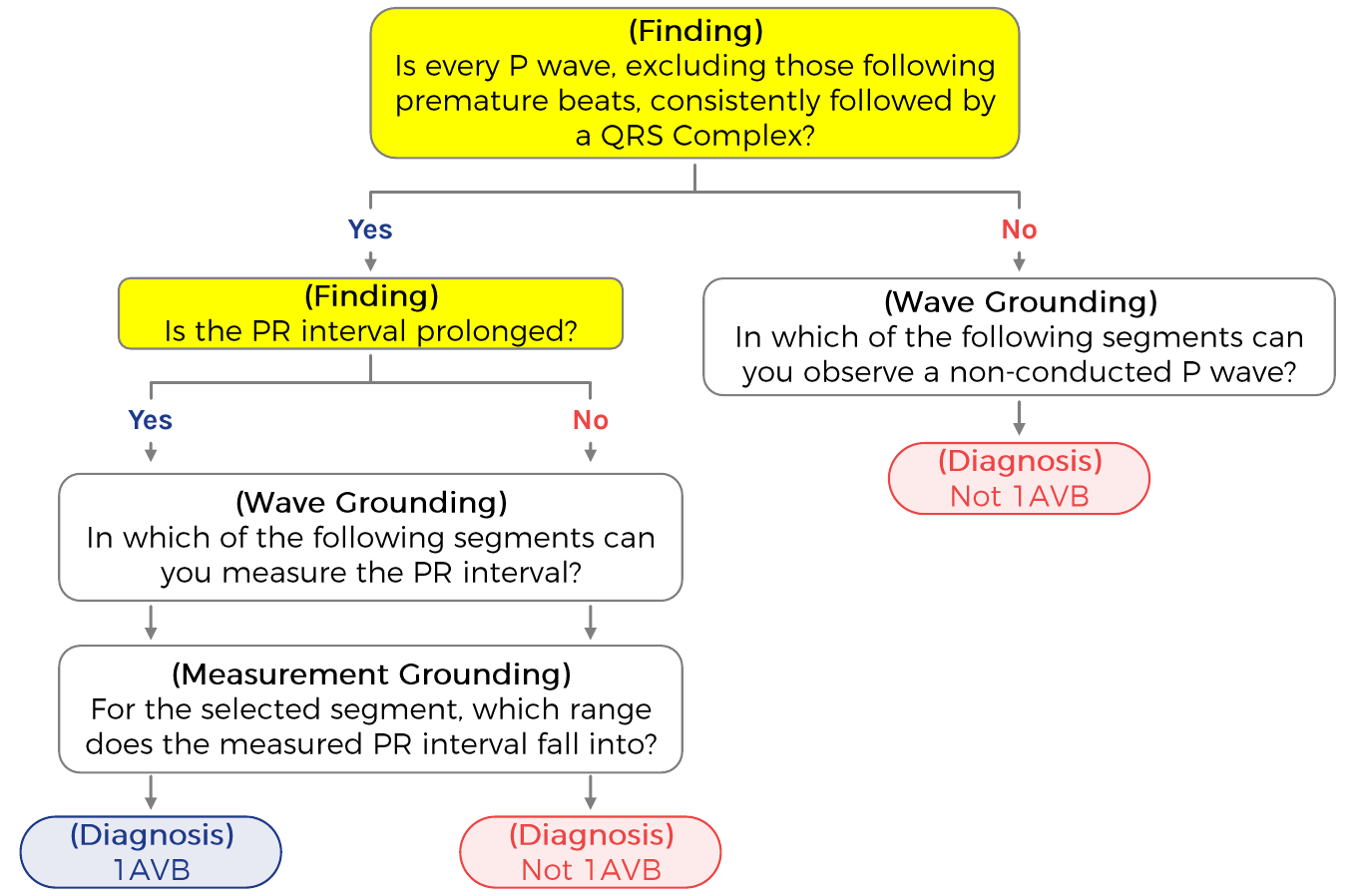}
    \caption{Logic diagram for First Degree AV Block (1AVB)}
    \label{fig:1avb}
\end{figure}

\begin{figure}[htbp]
    \centering
    \includegraphics[width=1.0\linewidth]{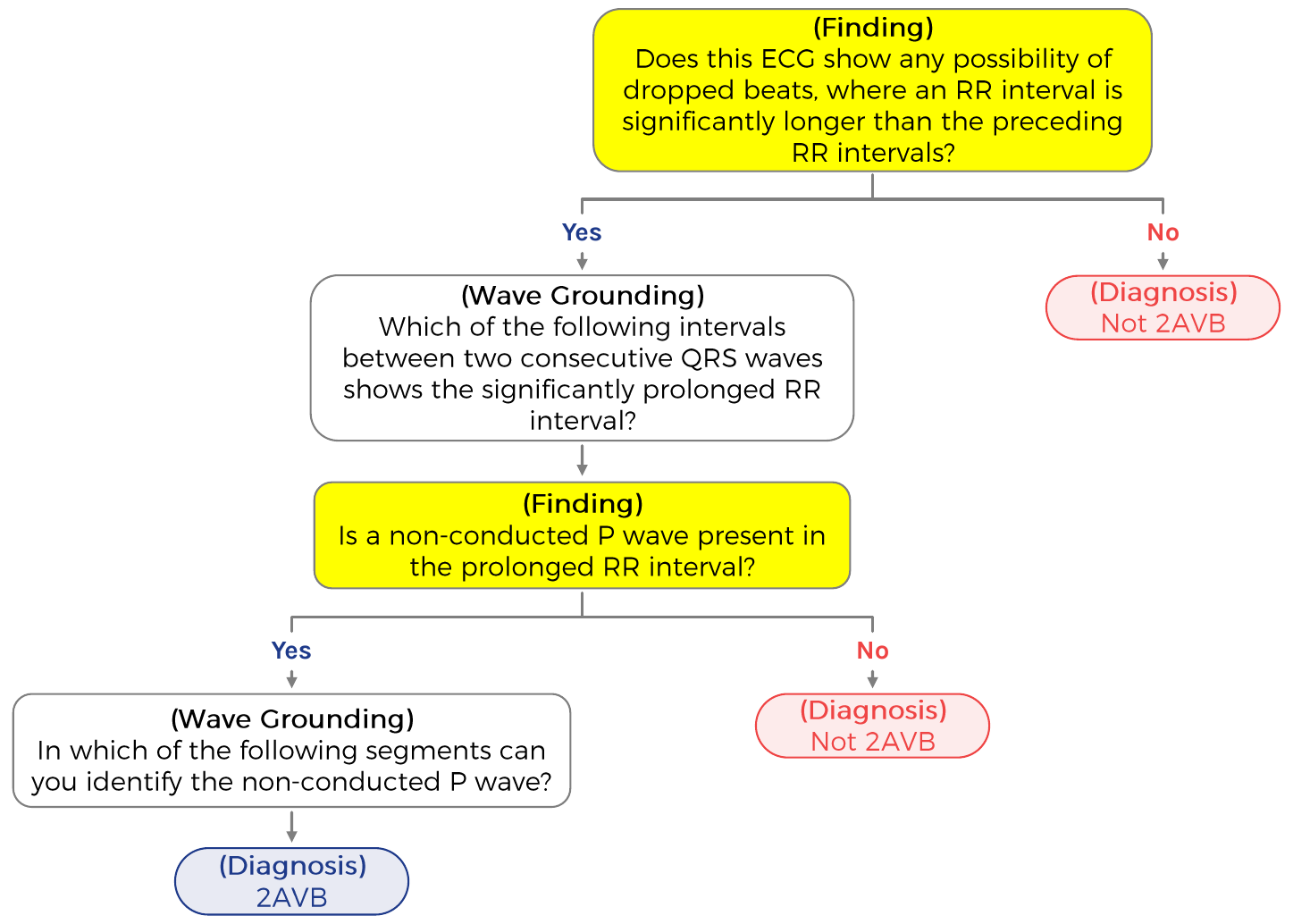}
    \caption{Logic diagram for Second Degree AV Block (2AVB)}
\end{figure}

\begin{figure}[htbp]
    \centering
    \includegraphics[width=1.0\linewidth]{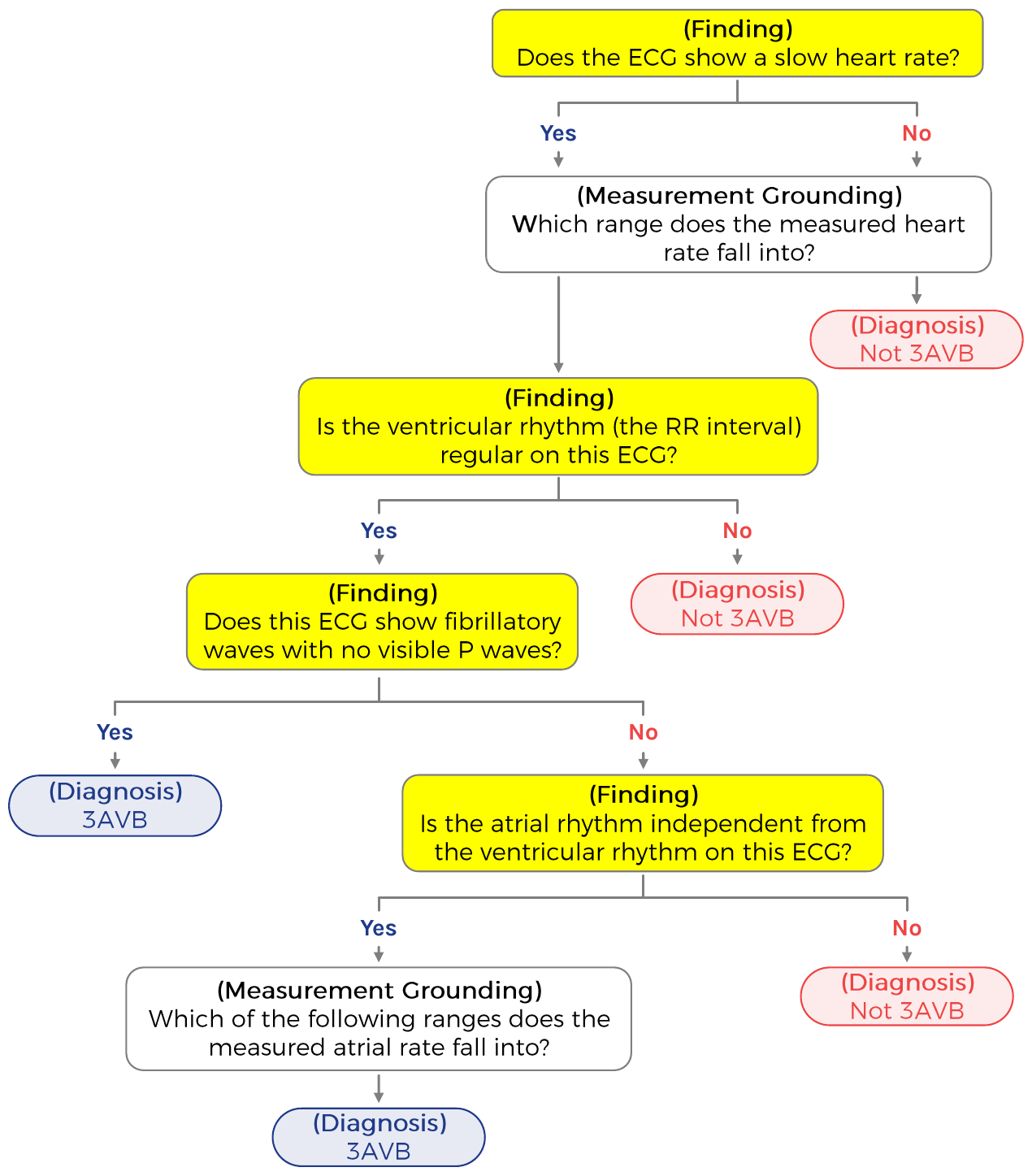}
    \caption{Logic diagram for Third Degree AV Block (3AVB)}
\end{figure}

\begin{figure}[htbp]
    \centering
    \includegraphics[width=1.0\linewidth]{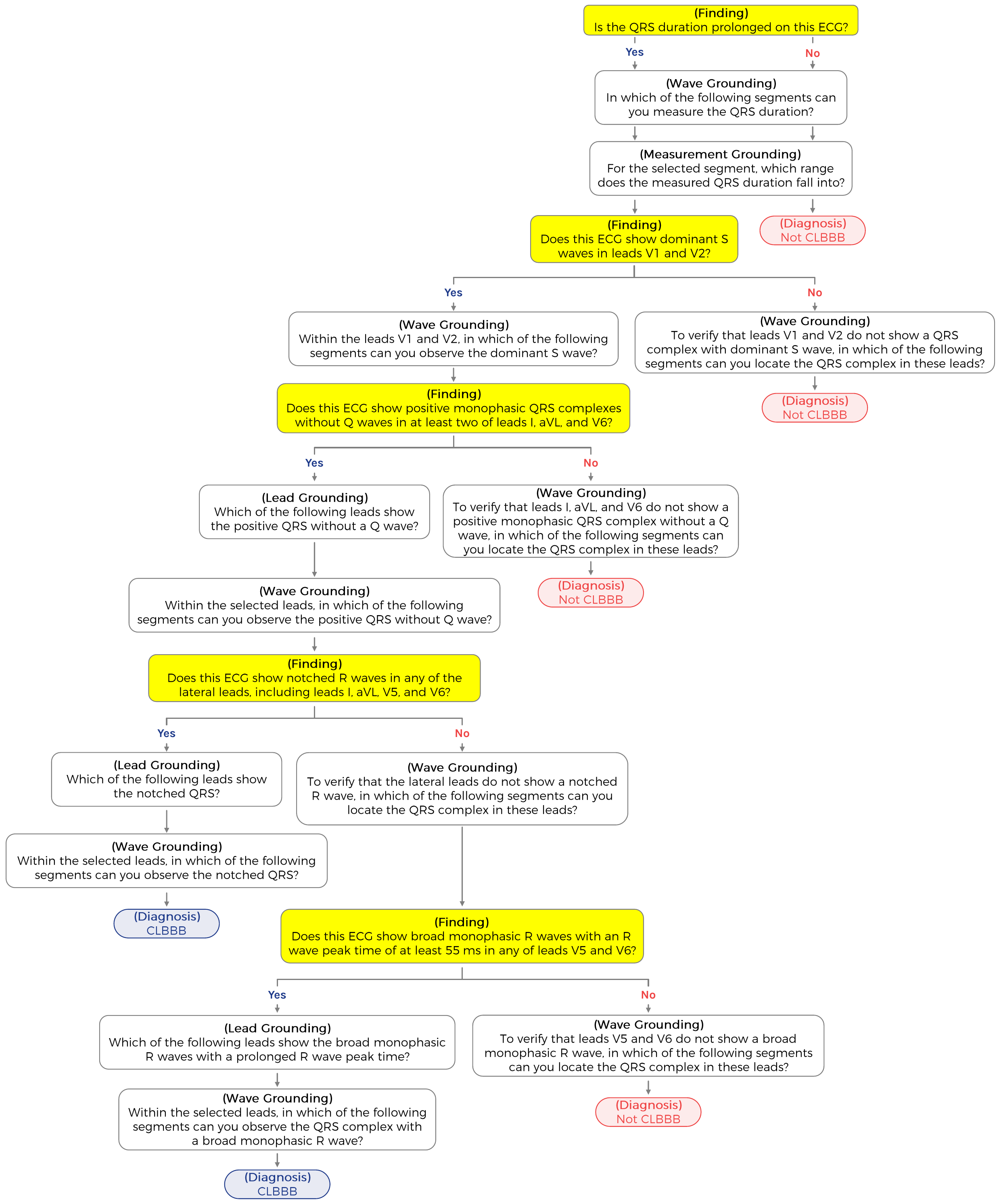}
    \caption{Logic diagram for Complete Left Bundle Branch Block (CLBBB)}
    \label{fig:diag_clbbb}
\end{figure}

\begin{figure}[htbp]
    \centering
    \includegraphics[width=1.0\linewidth]{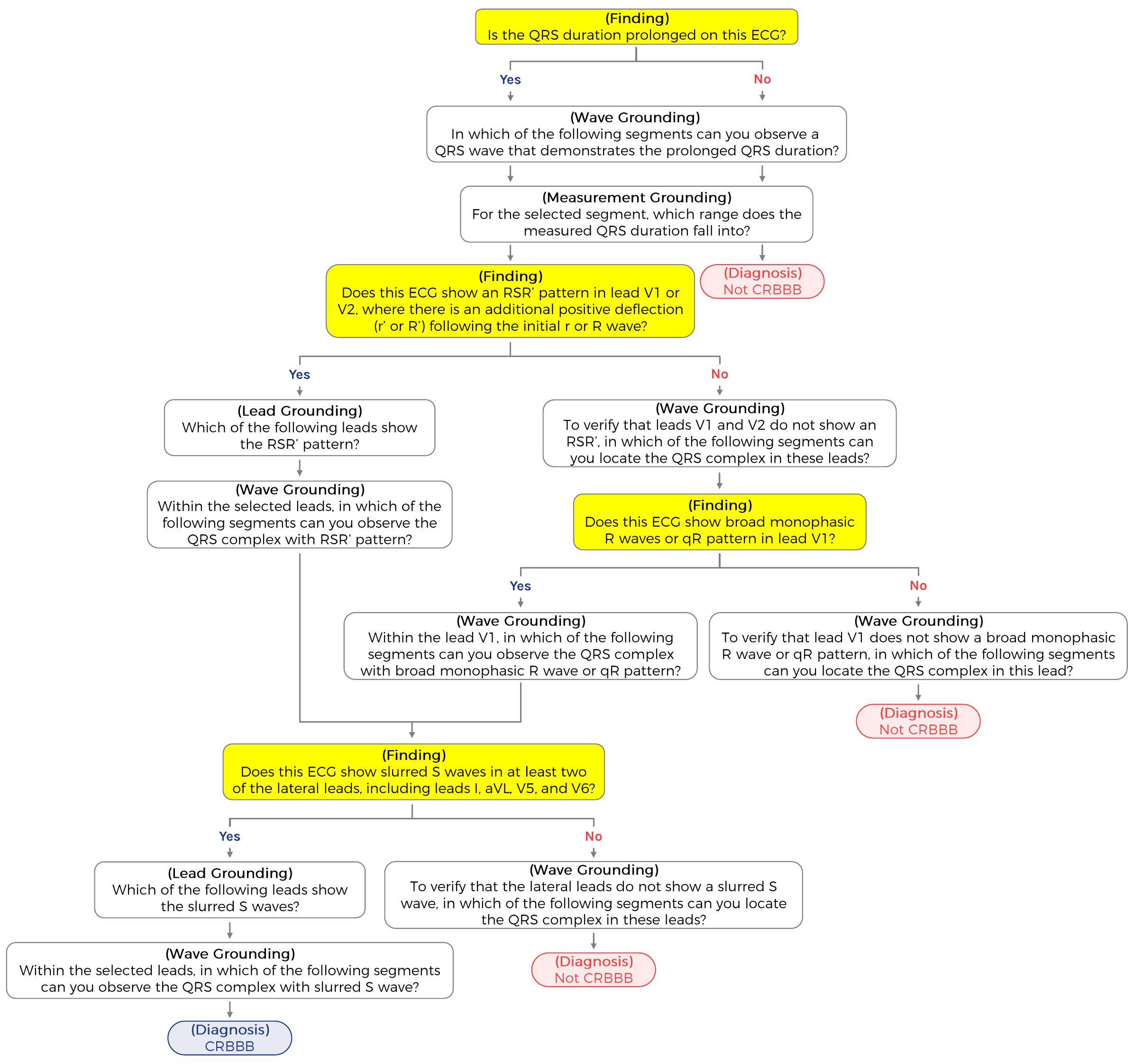}
    \caption{Logic diagram for Complete Right Bundle Branch Block (CRBBB)}
\end{figure}

\begin{figure}[htbp]
    \centering
    \includegraphics[width=1.0\linewidth]{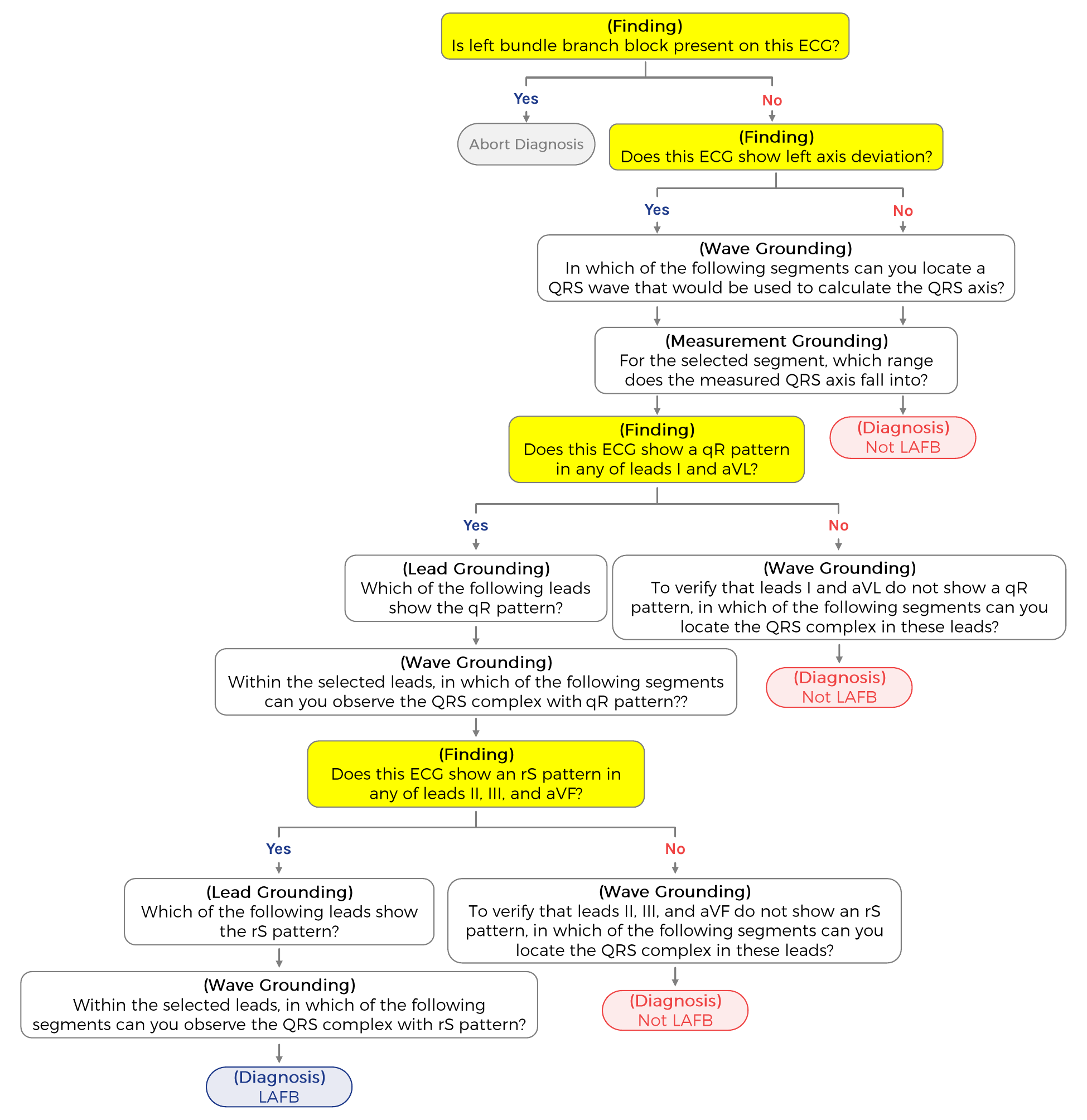}
    \caption{Logic diagram for Left Anterior Fascicular Block (LAFB)}
\end{figure}

\begin{figure}[htbp]
    \centering
    \includegraphics[width=1.0\linewidth]{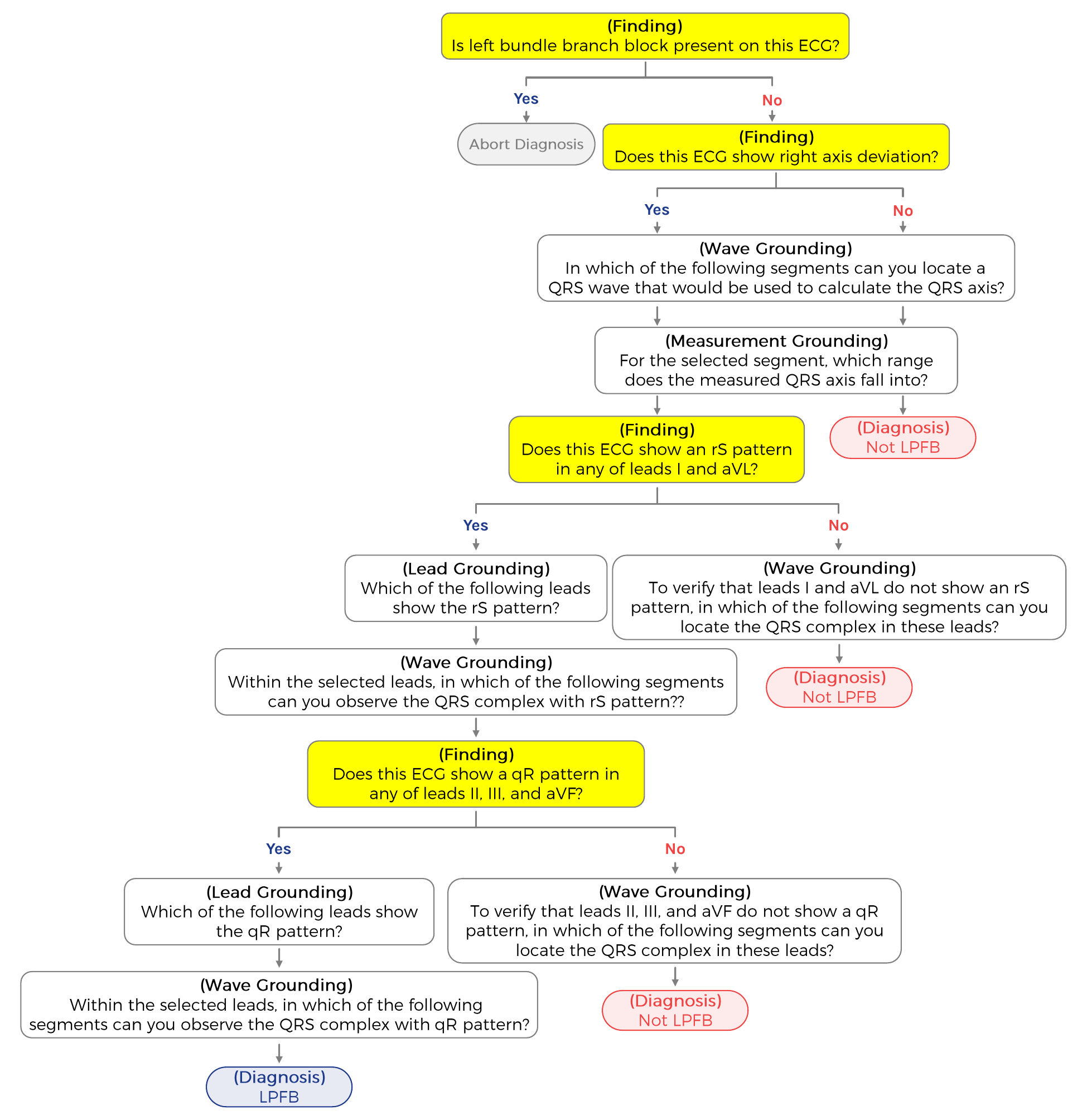}
    \caption{Logic diagram for Left Posterior Fascicular Block (LPFB)}
\end{figure}

\begin{figure}[htbp]
    \centering
    \includegraphics[width=1.0\linewidth]{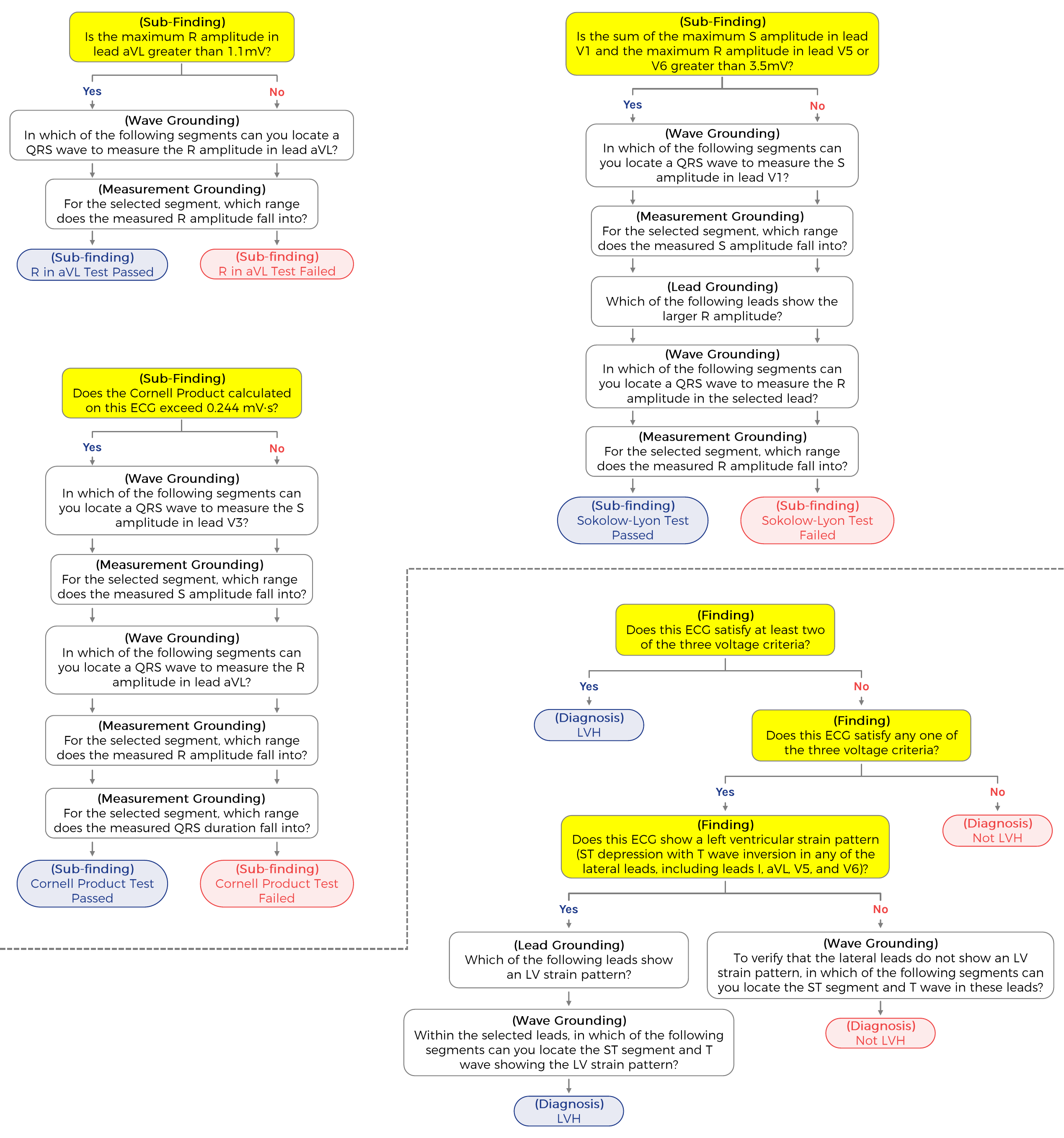}
    \caption{Logic diagram for Left Ventricular Hypertrophy (LVH)}
\end{figure}

\begin{figure}[htbp]
    \centering
    \includegraphics[width=1.0\linewidth]{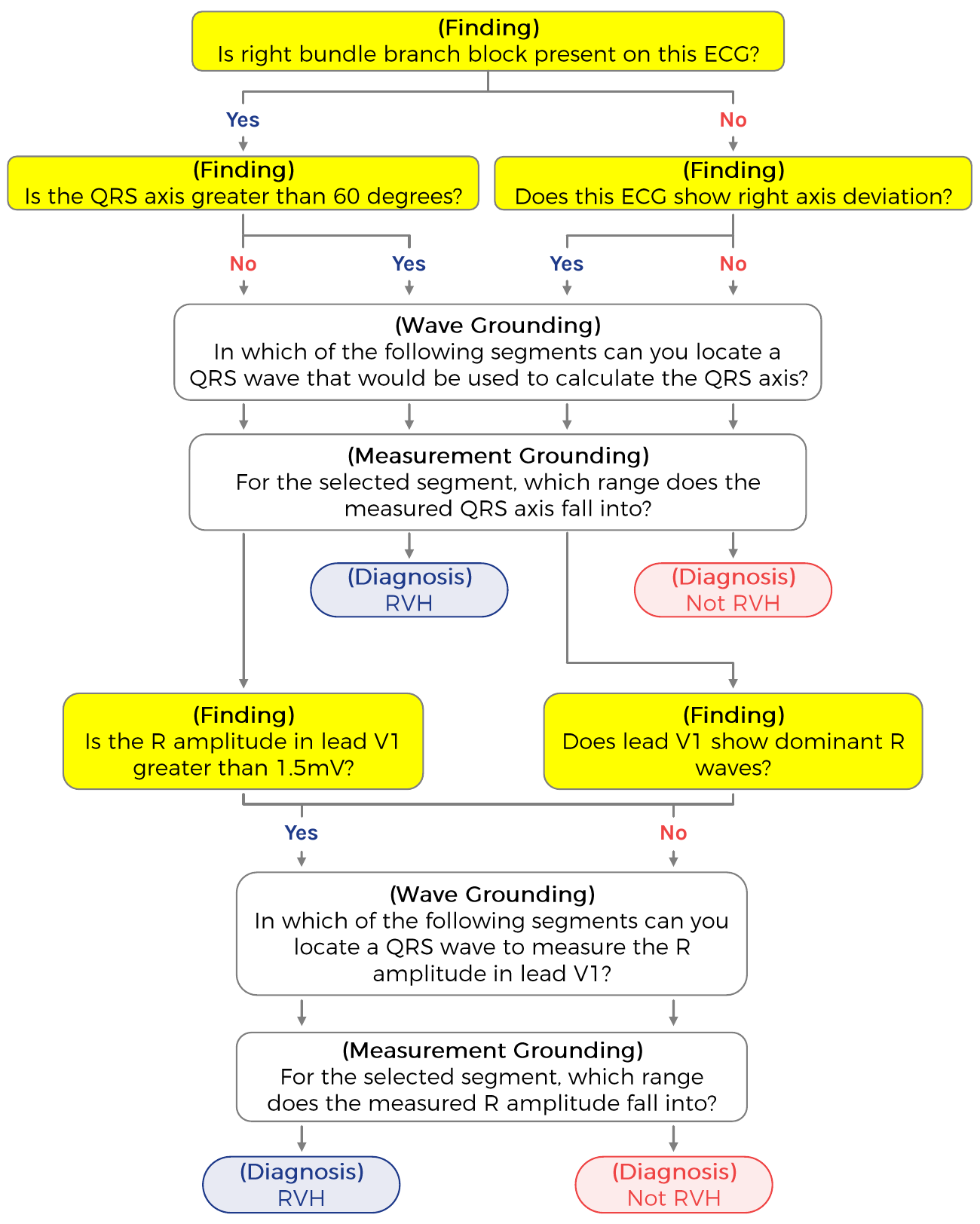}
    \caption{Logic diagram for Right Ventricular Hypertrophy (RVH)}
\end{figure}

\begin{figure}[htbp]
    \centering
    \includegraphics[width=1.0\linewidth]{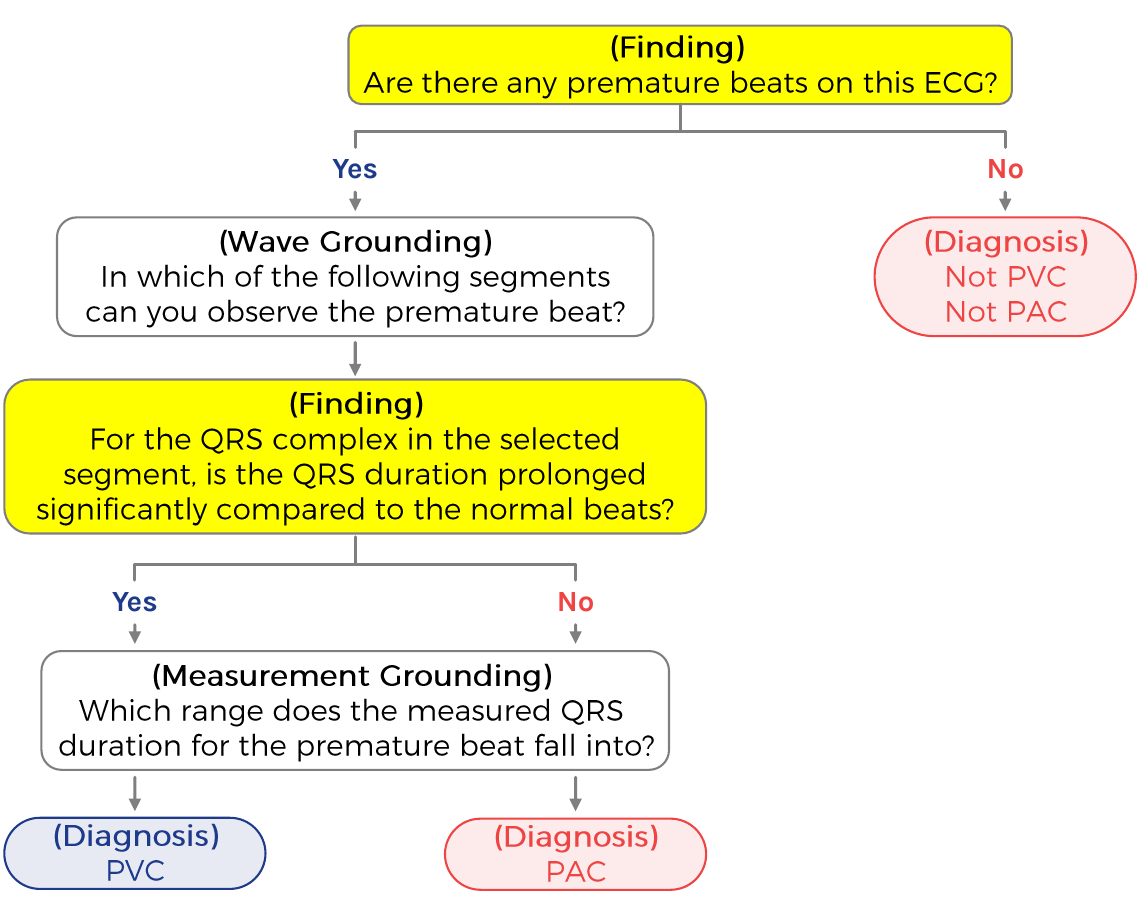}
    \caption{Logic diagram for Premature Atrial Complex (PAC) and Premature Ventricular Complex (PVC)}
\end{figure}

\begin{figure}[htbp]
    \centering
    \includegraphics[width=1.0\linewidth]{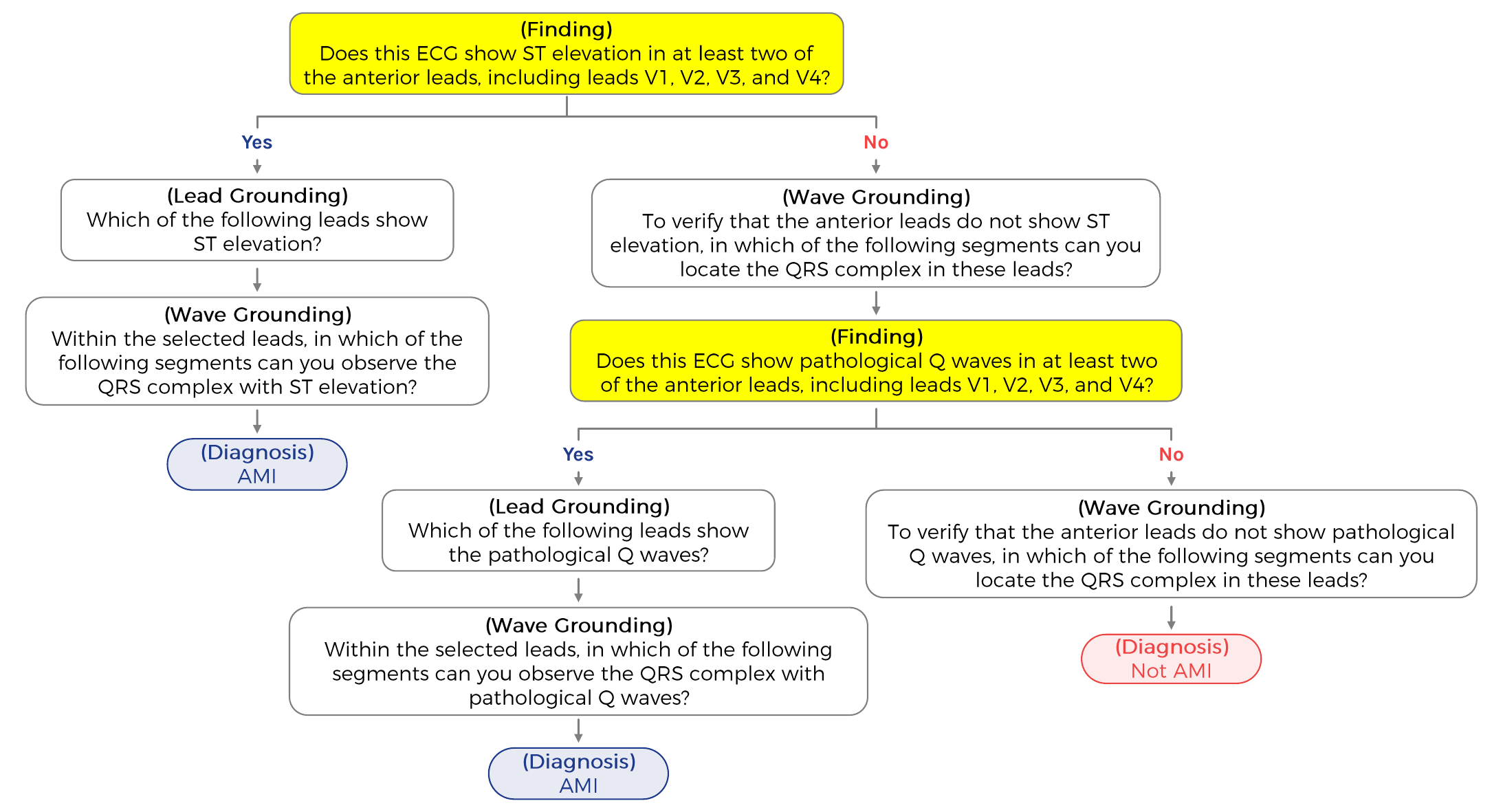}
    \caption{Logic diagram for Anterior Myocardial Infarction (AMI)}
\end{figure}

\begin{figure}[htbp]
    \centering
    \includegraphics[width=1.0\linewidth]{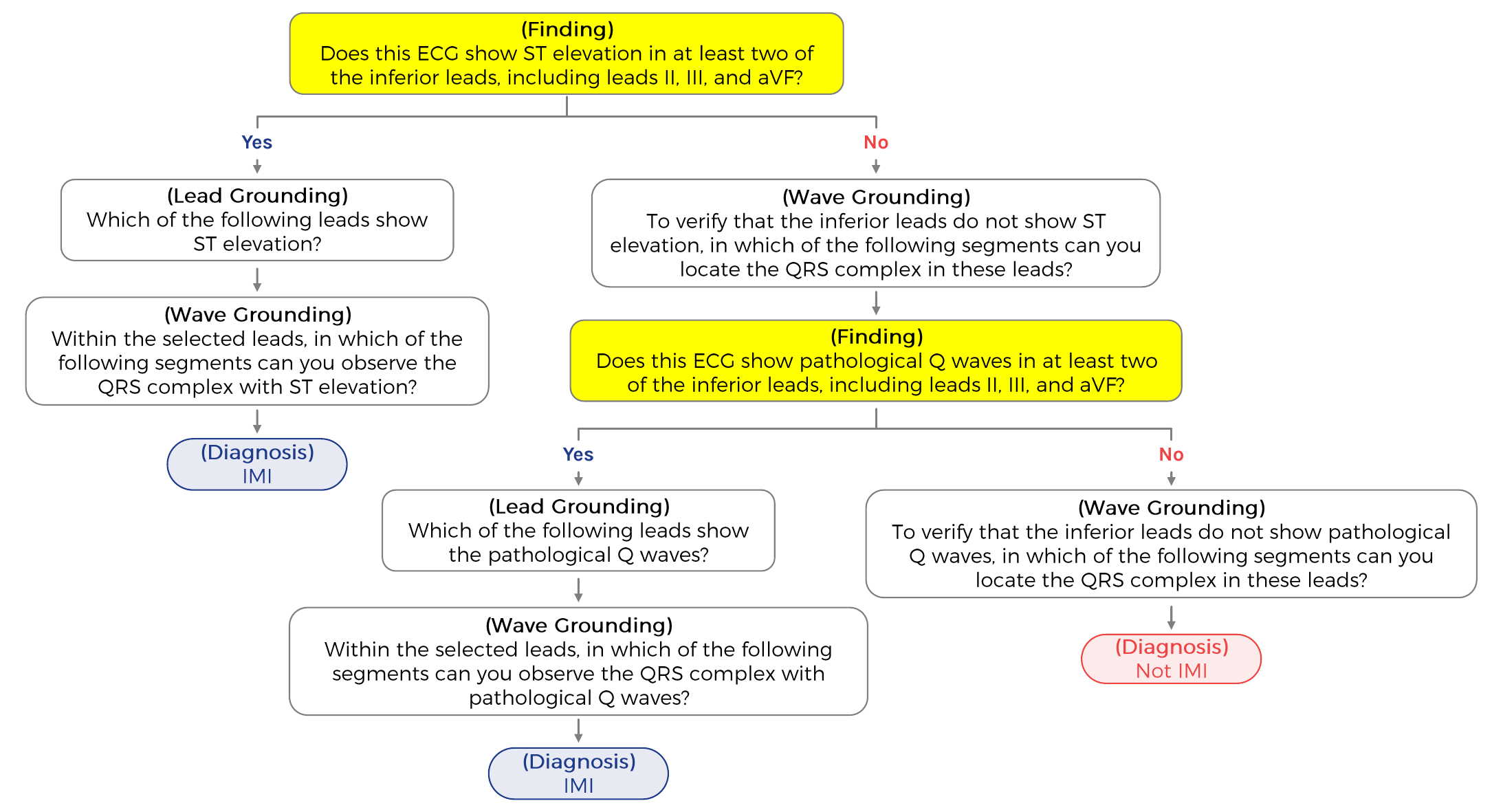}
    \caption{Logic diagram for Inferior Myocardial Infarction (IMI)}
\end{figure}

\begin{figure}[htbp]
    \centering
    \includegraphics[width=1.0\linewidth]{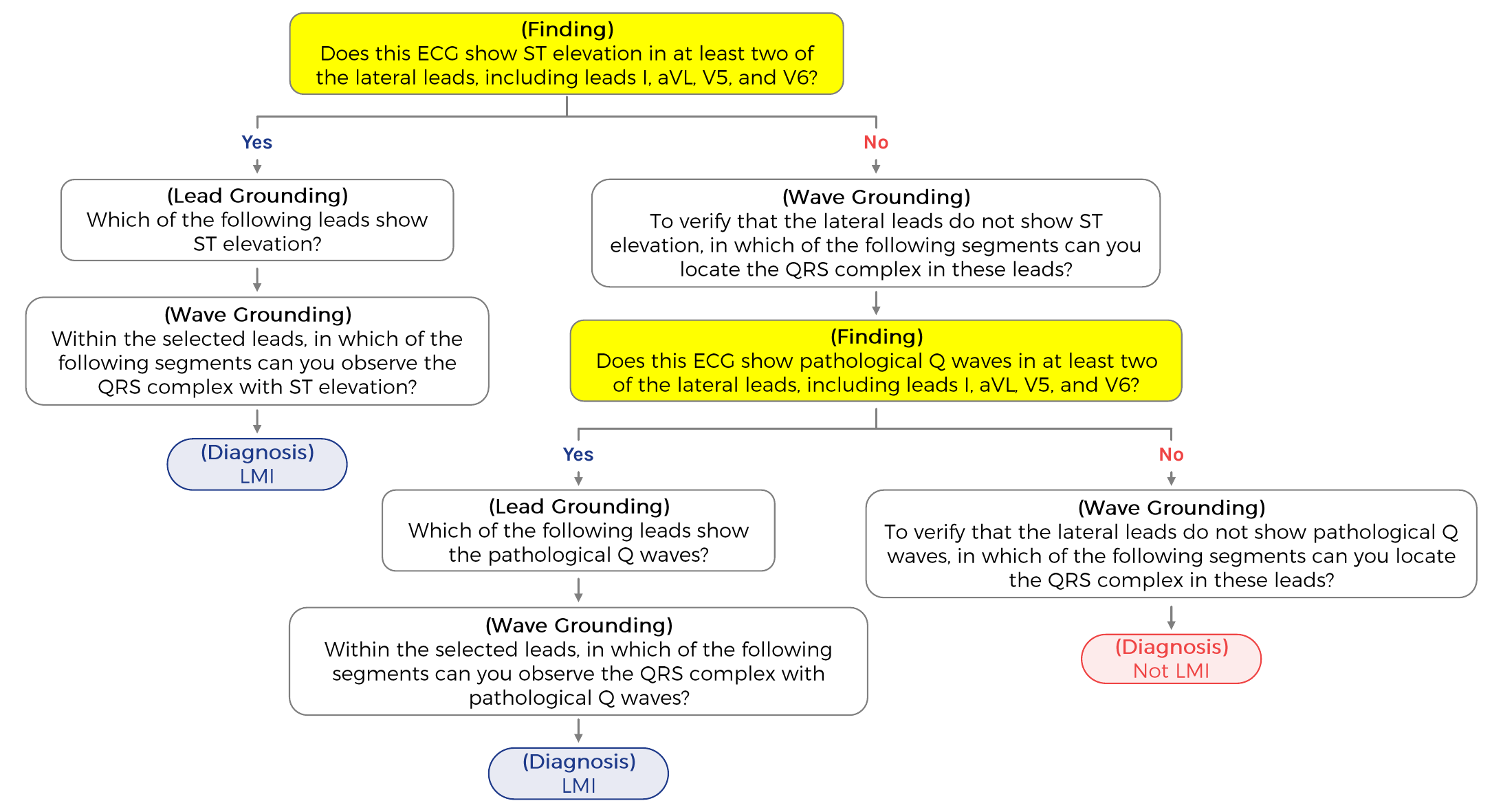}
    \caption{Logic diagram for Lateral Myocardial Infarction (LMI)}
\end{figure}

\begin{figure}[htbp]
    \centering
    \includegraphics[width=1.0\linewidth]{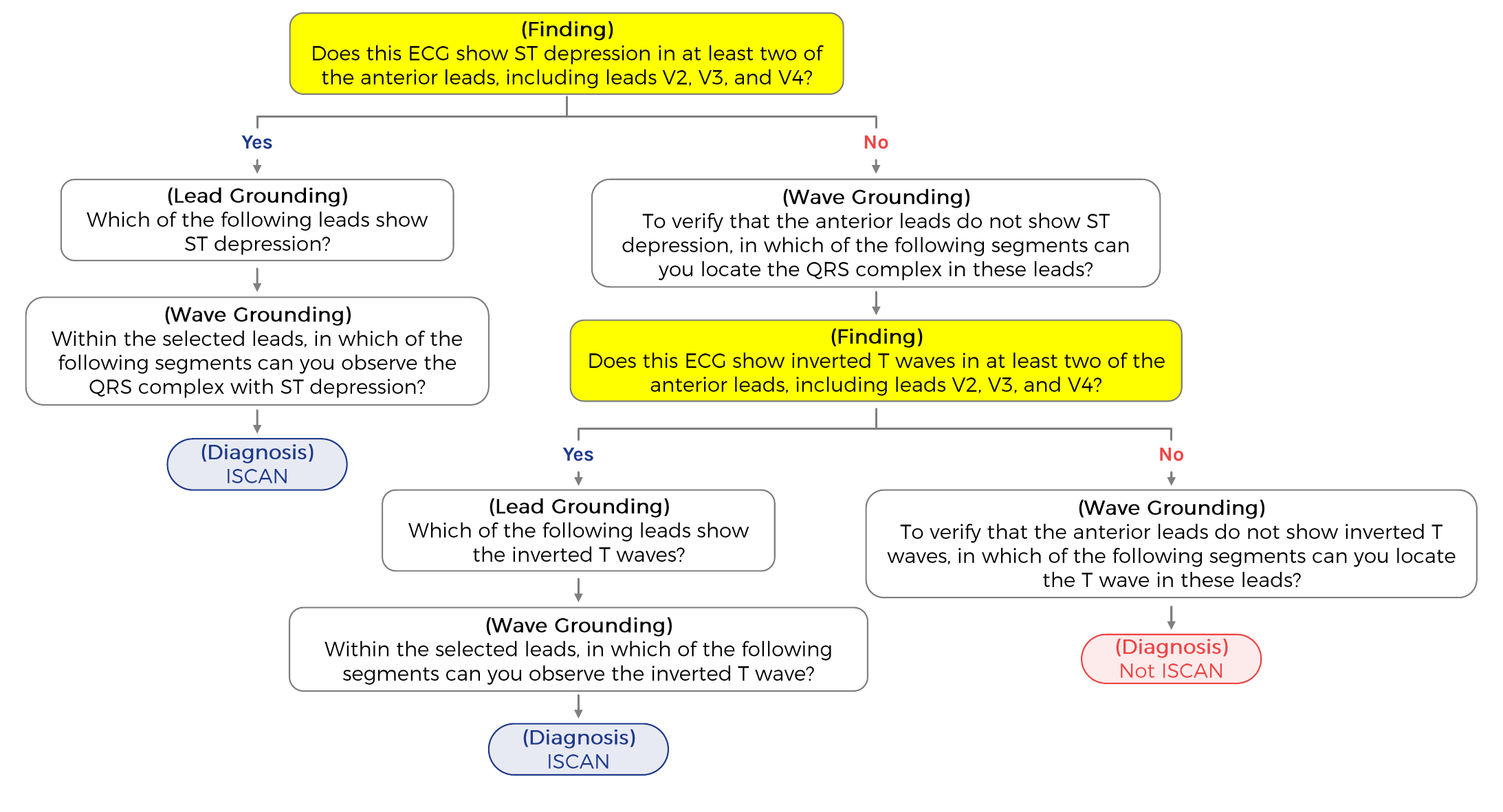}
    \caption{Logic diagram for Anterior Ischemia (ISCAN)}
\end{figure}

\begin{figure}[htbp]
    \centering
    \includegraphics[width=1.0\linewidth]{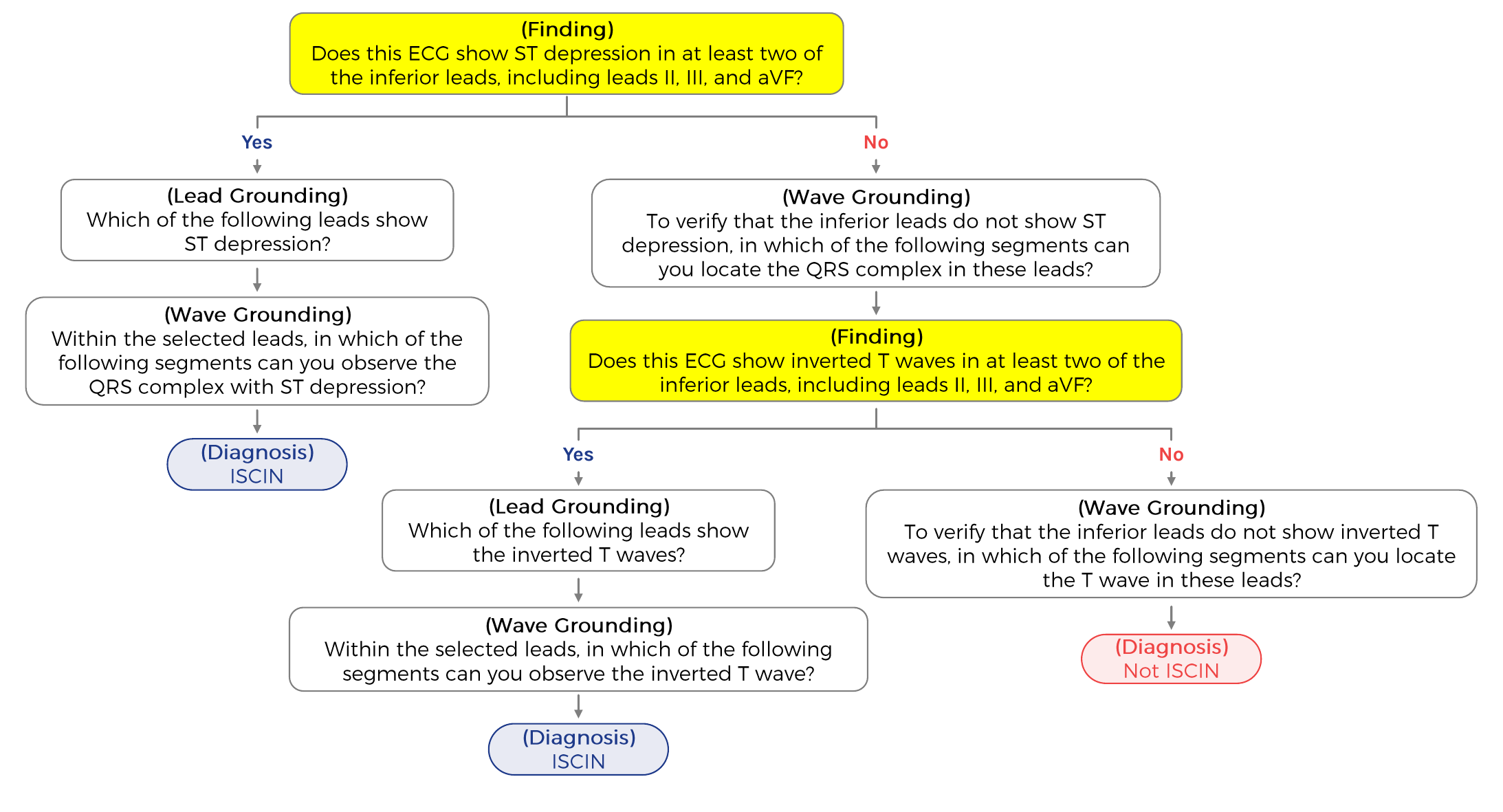}
    \caption{Logic diagram for Inferior Ischemia (ISCIN)}
\end{figure}

\begin{figure}[htbp]
    \centering
    \includegraphics[width=1.0\linewidth]{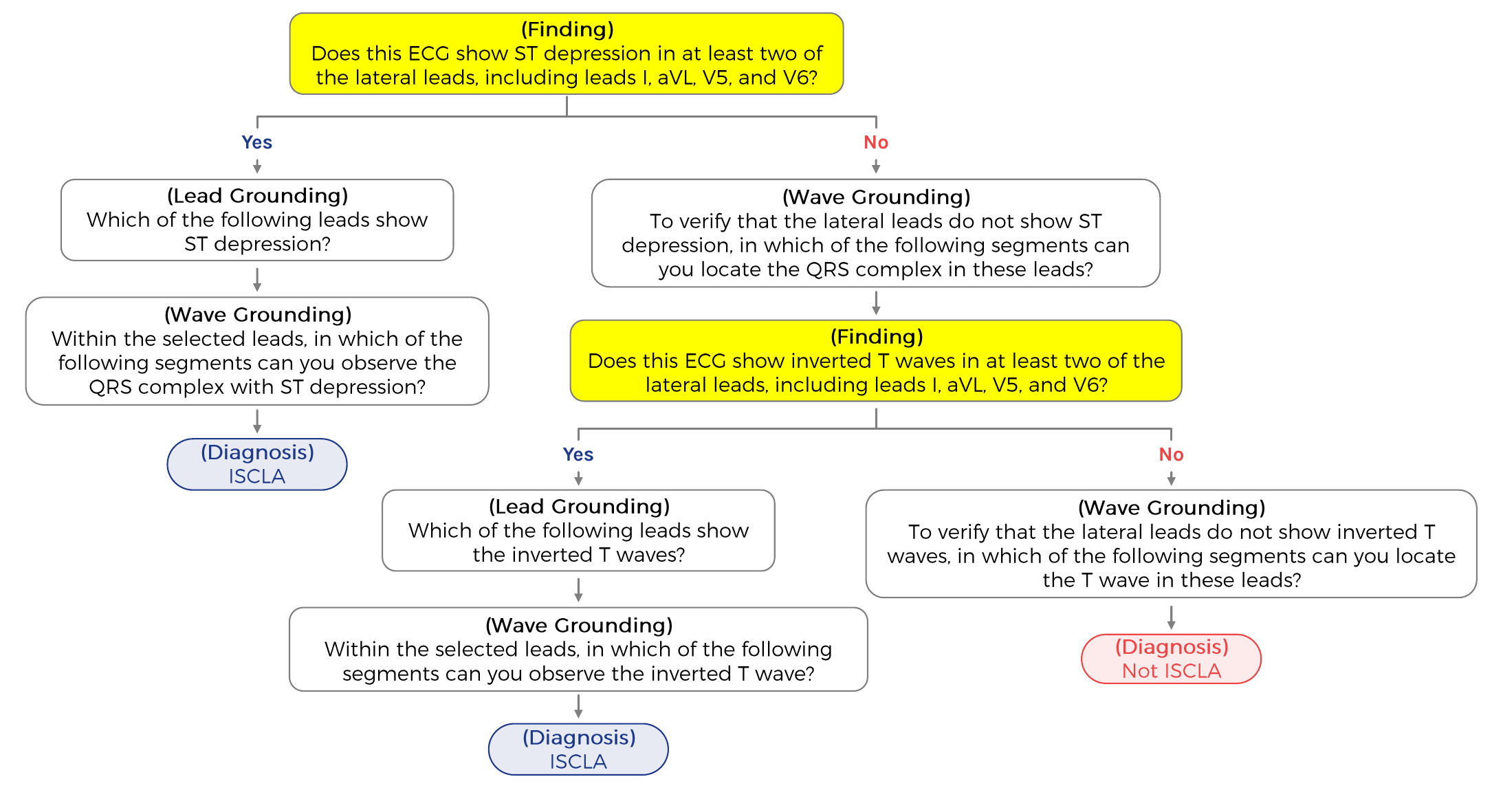}
    \caption{Logic diagram for Lateral Ischemia (ISCLA)}
    \label{fig:iscla}
\end{figure}

%%%%%%%%%%%%%%%%%%%%%%%%%%%%%%%%%%%%%%%%%%%%%%%%%%%%%%%%%%%%

\clearpage
\newpage
\section*{NeurIPS Paper Checklist}

\begin{enumerate}

\item {\bf Claims}
    \item[] Question: Do the main claims made in the abstract and introduction accurately reflect the paper's contributions and scope?
    \item[] Answer: \answerYes{} 
    \item[] Justification: The main claims in the abstract and introduction accurately reflect the paper's contributions and scope.

\item {\bf Limitations}
    \item[] Question: Does the paper discuss the limitations of the work performed by the authors?
    \item[] Answer: \answerYes{}
    \item[] Justification: The paper discusses the limitations in Section~\ref{sec:discussion}.

\item {\bf Theory assumptions and proofs}
    \item[] Question: For each theoretical result, does the paper provide the full set of assumptions and a complete (and correct) proof?
    \item[] Answer: \answerNA{}
    \item[] Justification: This paper does not include any theoretical results.

    \item {\bf Experimental result reproducibility}
    \item[] Question: Does the paper fully disclose all the information needed to reproduce the main experimental results of the paper to the extent that it affects the main claims and/or conclusions of the paper (regardless of whether the code and data are provided or not)?
    \item[] Answer: \answerYes{}
    \item[] Justification: The implementation details to reproduce the experiments, including the access points to the evaluated model and the detailed prompts used, are provided in Appendix~\ref{sec:app_impl_details}. We also provide source codes with detailed instructions in the repository: \url{https://github.com/Jwoo5/ecg-reasoning-benchmark}.

\item {\bf Open access to data and code}
    \item[] Question: Does the paper provide open access to the data and code, with sufficient instructions to faithfully reproduce the main experimental results, as described in supplemental material?
    \item[] Answer: \answerYes{}
    \item[] Justification: All the codes and data along with detailed scripts to reproduce the experimental results are publicly available in the provided repository link: \url{https://github.com/Jwoo5/ecg-reasoning-benchmark}.

\item {\bf Experimental setting/details}
    \item[] Question: Does the paper specify all the training and test details (e.g., data splits, hyperparameters, how they were chosen, type of optimizer) necessary to understand the results?
    \item[] Answer: \answerYes{}
    \item[] Justification: The details are provided in Appendix~\ref{sec:app_impl_details}, as well as with the provided source codes.

\item {\bf Experiment statistical significance}
    \item[] Question: Does the paper report error bars suitably and correctly defined or other appropriate information about the statistical significance of the experiments?
    \item[] Answer: \answerNo{}
    \item[] Justification: We did not include error bars because it was not feasible to run experiments multiple times given the computational burden of repeated inference, as well as the substantial API costs required for evaluating proprietary models.

\item {\bf Experiments compute resources}
    \item[] Question: For each experiment, does the paper provide sufficient information on the computer resources (type of compute workers, memory, time of execution) needed to reproduce the experiments?
    \item[] Answer: \answerYes{}
    \item[] Justification: We provide these details in Appendix~\ref{sec:app_impl_details}.
    
\item {\bf Code of ethics}
    \item[] Question: Does the research conducted in the paper conform, in every respect, with the NeurIPS Code of Ethics \url{https://neurips.cc/public/EthicsGuidelines}?
    \item[] Answer: \answerYes{}
    \item[] Justification: This paper conforms to the NeurIPS Code of Ethics.

\item {\bf Broader impacts}
    \item[] Question: Does the paper discuss both potential positive societal impacts and negative societal impacts of the work performed?
    \item[] Answer: \answerNA{}
    \item[] Justification: The proposed benchmark dataset is entirely based on MIMIC-IV-ECG and PTB-XL, which have been de-identified and are publicly available. Given that, we do not anticipate any negative societal impacts or privacy concerns related to our work.
    
\item {\bf Safeguards}
    \item[] Question: Does the paper describe safeguards that have been put in place for responsible release of data or models that have a high risk for misuse (e.g., pre-trained language models, image generators, or scraped datasets)?
    \item[] Answer: \answerNA{}
    \item[] Justification: Since the proposed benchmark dataset is entirely based on de-identified public datasets, MIMIC-IV-ECG and PTB-XL, our work does not pose a high risk for misuse, and thus no additional safeguards are required.

\item {\bf Licenses for existing assets}
    \item[] Question: Are the creators or original owners of assets (e.g., code, data, models), used in the paper, properly credited and are the license and terms of use explicitly mentioned and properly respected?
    \item[] Answer: \answerYes{}
    \item[] Justification: All assets used in the paper are publicly available and properly cited.

\item {\bf New assets}
    \item[] Question: Are new assets introduced in the paper well documented and is the documentation provided alongside the assets?
    \item[] Answer: \answerYes{}
    \item[] Justification: We provide the detailed documentation alongside the code.

\item {\bf Crowdsourcing and research with human subjects}
    \item[] Question: For crowdsourcing experiments and research with human subjects, does the paper include the full text of instructions given to participants and screenshots, if applicable, as well as details about compensation (if any)? 
    \item[] Answer: \answerNA{}
    \item[] Justification: The paper does not involve crowdsourcing nor research with human subjects.

\item {\bf Institutional review board (IRB) approvals or equivalent for research with human subjects}
    \item[] Question: Does the paper describe potential risks incurred by study participants, whether such risks were disclosed to the subjects, and whether Institutional Review Board (IRB) approvals (or an equivalent approval/review based on the requirements of your country or institution) were obtained?
    \item[] Answer: \answerNA{}
    \item[] Justification: The paper does not involve crowdsourcing nor research with human subjects.

\item {\bf Declaration of LLM usage}
    \item[] Question: Does the paper describe the usage of LLMs if it is an important, original, or non-standard component of the core methods in this research? Note that if the LLM is used only for writing, editing, or formatting purposes and does \emph{not} impact the core methodology, scientific rigor, or originality of the research, declaration is not required.
    %this research? 
    \item[] Answer: \answerNA{}
    \item[] Justification: The core method development in this research does not involve LLMs as any important, original, or non-standard components.

\end{enumerate}

\end{document}